%% file: sn-article.tex
\documentclass[10pt]{article}

\usepackage{graphicx}
\usepackage{flafter}
\usepackage{multirow}
\usepackage{amsmath,amssymb,amsfonts}
\usepackage{amsthm}
\usepackage{mathrsfs}
\usepackage{textcomp}
\usepackage{booktabs}
\usepackage{algorithm}
\usepackage{algorithmicx}
\usepackage{algpseudocode}
\usepackage{listings}
\usepackage{array}
\usepackage{tabularx}
\usepackage{multicol}
\usepackage{acronym}
\usepackage{auin-arxiv}
\usepackage{colortbl}
\usepackage[numbers,sort&compress]{natbib}
\usepackage{xurl}
\usepackage{fontawesome5}
\usepackage[colorlinks=true,linkcolor=AUINBlue,citecolor=AUINBlue,
                        urlcolor=AUINBlue,anchorcolor=AUINNavy]{hyperref}

\theoremstyle{plain}

\theoremstyle{definition}

\newtheorem{definition}{Definition}
\theoremstyle{remark}

\newacro{3D-CRNN}{3D Convolutional Recurrent Neural Network}
\newacro{AEB}{Automated Emergency Braking}
\newacro{ASR}{Artifact Subspace Reconstruction}
\newacro{AV}{Autonomous Vehicle}
\newacro{BA}{Balanced Accuracy}
\newacro{BCI}{Brain--Computer Interface}
\newacro{CNN}{Convolutional Neural Network}
\newacro{CSP}{Common Spatial Pattern}
\newacro{DI}{Danger Identification}
\newacro{EEG}{Electroencephalogram}
\newacro{FBCSP}{Filter Bank Common Spatial Pattern}
\newacro{FIR}{Finite Impulse Response}
\newacro{FLOP}{Floating-Point Operation}
\newacro{fMRI}{Functional Magnetic Resonance Imaging}
\newacro{fNIRS}{Functional Near-Infrared Spectroscopy}
\newacro{GRU}{Gated Recurrent Unit}
\newacro{ICA}{Independent Component Analysis}
\newacro{LDA}{Linear Discriminant Analysis}
\newacro{LOSO}{Leave-One-Subject-Out}
\newacro{LSTM}{Long Short-Term Memory}
\newacro{MEG}{Magnetoencephalography}
\newacro{PCM}{Passenger Cognitive Model}
\newacro{PEDS}{Passenger EEG Decoding Strategy}
\newacro{PSNR}{Peak Signal-to-Noise Ratio}
\newacro{RP}{Risk Prediction}
\newacro{RSL}{Risk-aware Sequential Labeling}
\newacro{RSVP}{Rapid Serial Visual Presentation}
\newacro{SOTIF}{Safety of the Intended Functionality}
\newacro{SVM}{Support Vector Machine}
\newacro{VTD}{Virtual Test Drive}

\hypersetup{
    pdftitle={EEG-Driven Decoding Framework for Passenger Hazard Perception in Highly Automated Vehicles},
    pdfauthor={Yingkai Yang, Ashton Yu Xuan Tan, Bowen Li, Xiaorong Gao, Sifa Zheng, Jianqiang Wang, Xinyu Gu, Yang Zhao, Yuxin Zhang, Sharon X. Huang, Tania Stathaki, Jun Li, Hong Wang},
    pdfsubject={Accepted manuscript for Automotive Innovation},
    pdfkeywords={Autonomous driving, passenger cognition, accident prevention, human factors, electroencephalogram}
}
\preprintstatus{Accepted manuscript, Automotive Innovation}
\preprintshorttitle{EEG-Driven Passenger Hazard Perception}

\title{EEG-Driven Decoding Framework for Passenger Hazard Perception in Highly Automated Vehicles}

\author{%
    {\large\sffamily\bfseries
        Yingkai Yang\textsuperscript{1,2,\textdagger}\quad
        Ashton Yu Xuan Tan\textsuperscript{1,\textdagger}\quad
        Bowen Li\textsuperscript{3}\quad
        Xiaorong Gao\textsuperscript{3}\\[0.25em]
        Sifa Zheng\textsuperscript{1}\quad
        Jianqiang Wang\textsuperscript{1}\quad
        Xinyu Gu\textsuperscript{1}\quad
        Yang Zhao\textsuperscript{4}\quad
        Yuxin Zhang\textsuperscript{5}\\[0.25em]
        Sharon X. Huang\textsuperscript{6}\quad
        Tania Stathaki\textsuperscript{2}\quad
        Jun Li\textsuperscript{1}\quad
        Hong Wang\textsuperscript{1,*}\\[0.75em]
    }
    {\small
        \textsuperscript{1}School of Vehicle and Mobility, Tsinghua University, Beijing 100084, China\\
        \textsuperscript{2}Department of Electrical and Electronic Engineering, Imperial College London, London SW7 2AZ, United Kingdom\\
        \textsuperscript{3}School of Medicine, Tsinghua University, Beijing 100084, China\\
        \textsuperscript{4}School of Automation Engineering, University of Electronic Science and Technology of China, Chengdu 611731, China\\
        \textsuperscript{5}State Key Laboratory of Automotive Simulation and Control, Jilin University, Changchun 130025, China\\
        \textsuperscript{6}College of Information Sciences and Technology, Pennsylvania State University, Pennsylvania 16802, United States of America\\[0.55em]
    }
    {\footnotesize
        \textsuperscript{\textdagger}These authors contributed equally to this work.\quad
        \textsuperscript{*}Corresponding author: \href{mailto:hong_wang@tsinghua.edu.cn}{\nolinkurl{hong_wang@tsinghua.edu.cn}}.\\[0.35em]
    }
    {\scriptsize
        \begin{minipage}{0.98\textwidth}\centering
        \textit{Author emails:}
        \href{mailto:yingkai.yang19@imperial.ac.uk}{\nolinkurl{yingkai.yang19@imperial.ac.uk}},
        \href{mailto:yuxuan-c23@mails.tsinghua.edu.cn}{\nolinkurl{yuxuan-c23@mails.tsinghua.edu.cn}},
        \href{mailto:bowenlieeg@tsinghua.edu.cn}{\nolinkurl{bowenlieeg@tsinghua.edu.cn}},
        \href{mailto:gxr-dea@tsinghua.edu.cn}{\nolinkurl{gxr-dea@tsinghua.edu.cn}},
        \href{mailto:zsf@tsinghua.edu.cn}{\nolinkurl{zsf@tsinghua.edu.cn}},
        \href{mailto:wjqlws@tsinghua.edu.cn}{\nolinkurl{wjqlws@tsinghua.edu.cn}},
        \href{mailto:guxinyu_98@163.com}{\nolinkurl{guxinyu_98@163.com}},
        \href{mailto:zyuestc@126.com}{\nolinkurl{zyuestc@126.com}},
        \href{mailto:yuxinzhang@jlu.edu.cn}{\nolinkurl{yuxinzhang@jlu.edu.cn}},
        \href{mailto:suh972@psu.edu}{\nolinkurl{suh972@psu.edu}},
        \href{mailto:t.stathaki@imperial.ac.uk}{\nolinkurl{t.stathaki@imperial.ac.uk}},
        \href{mailto:lijun1958@tsinghua.edu.cn}{\nolinkurl{lijun1958@tsinghua.edu.cn}}, and
        \href{mailto:hong_wang@tsinghua.edu.cn}{\nolinkurl{hong_wang@tsinghua.edu.cn}}.
        \end{minipage}
    }
    \\[0.5em]
    {\small\sffamily\centering
        \href{https://ieee-dataport.org/documents/passengereeg-eeg-dataset-passenger-hazard-perception-avs}{\faGlobe\enspace PassengerEEG}
        \qquad
        \href{https://github.com/SOTIF-AVLab/EEG2023}{\faGithub\enspace GitHub}
    }
}
\date{}

\begin{document}
\maketitle

\begin{abstract}
Reliable risk assessment remains a central challenge for \acfp{AV}. Despite advances in automation, passenger cognition provides a non-intrusive auxiliary signal that improves both objective and perceived safety without requiring active human intervention. We introduce an \acf{EEG}-based \acf{BCI} that decodes passenger neural responses for both \acf{RP} and \acf{DI}, explicitly modeling humans as passengers to match real-world \ac{AV} use. To achieve this, we propose the \acf{PCM}, \acf{RSL}, and the \acf{PEDS}, which integrates a \acf{3D-CRNN} model for joint EEG decoding. Experimental results show that \ac{3D-CRNN} achieves a \acf{BA} of \(95.3\% \pm 2.7\%\) in \ac{RP} and improves single-subject \ac{DI} from \(80.9\% \pm 3.9\%\) to \(85.0\% \pm 3.2\%\) with \ac{RSL}. Event-wise analyses further show that \ac{3D-CRNN} consistently outperforms other models across different event types in \ac{RP} and \ac{DI}. In generalization experiments, \ac{3D-CRNN} achieves \(77.0\% \pm 5.3\%\) \ac{BA} in cross-session \ac{DI} and \(77.4\% \pm 1.1\%\) \ac{BA} on seen subjects in cross-subject evaluation, while maintaining a \(64.9\% \pm 8.5\%\) \ac{BA} on unseen subjects, demonstrating promising generalizability and transferability across both intra-subject and inter-subject variability. These findings establish an \acf{EEG} decoding framework for \ac{AV} passenger hazard perception and suggest that passenger cognitive signals can provide auxiliary supervision for future \ac{AV} decision-making and \acf{SOTIF} support.
\end{abstract}


\clearpage
\section*{List of Acronyms}
\noindent\begin{minipage}{\textwidth}
\begin{multicols}{2}
\small
\setlength{\parskip}{0pt}
\begin{acronym}[3D-CRNN]
    \acro{3D-CRNN}{3D Convolutional Recurrent Neural Network}
    \acro{AEB}{Automated Emergency Braking}
    \acro{ASR}{Artifact Subspace Reconstruction}
    \acro{AV}{Autonomous Vehicle}
    \acro{BA}{Balanced Accuracy}
    \acro{BCI}{Brain--Computer Interface}
    \acro{CNN}{Convolutional Neural Network}
    \acro{CSP}{Common Spatial Pattern}
    \acro{DI}{Danger Identification}
    \acro{EEG}{Electroencephalogram}
    \acro{FBCSP}{Filter Bank Common Spatial Pattern}
    \acro{FIR}{Finite Impulse Response}
    \acro{FLOP}{Floating-Point Operation}
    \acro{fMRI}{Functional Magnetic Resonance Imaging}
    \acro{fNIRS}{Functional Near-Infrared Spectroscopy}
    \acro{GRU}{Gated Recurrent Unit}
    \acro{ICA}{Independent Component Analysis}
    \acro{LDA}{Linear Discriminant Analysis}
    \acro{LOSO}{Leave-One-Subject-Out}
    \acro{LSTM}{Long Short-Term Memory}
    \acro{MEG}{Magnetoencephalography}
    \acro{PCM}{Passenger Cognitive Model}
    \acro{PEDS}{Passenger EEG Decoding Strategy}
    \acro{PSNR}{Peak Signal-to-Noise Ratio}
    \acro{RP}{Risk Prediction}
    \acro{RSL}{Risk-aware Sequential Labeling}
    \acro{RSVP}{Rapid Serial Visual Presentation}
    \acro{SOTIF}{Safety of the Intended Functionality}
    \acro{SVM}{Support Vector Machine}
    \acro{VTD}{Virtual Test Drive}
\end{acronym}
\end{multicols}
\end{minipage}

\section{Introduction}
\acresetall{}
In recent years, the occurrence of \ac{SOTIF} accidents has placed critical safety issues associated with \acp{AV} in the spotlight~\cite{tesla2016,uber2018,tesla2020,Yu2024,bib1}. Two notable examples of such incidents involve an automated vehicle colliding with an overturned white truck and another high-level automated vehicle failing to avoid a collision with an illegally crossing cyclist~\cite{tesla2020,uber2018}. A key factor in these accidents is the inadequacy of data-driven algorithms, which struggle with unforeseen scenarios due to limited training data coverage and the challenges of operating in unpredictable, boundaryless environments.

More broadly, recent work on human-intelligence-augmented AI has emphasized the value of incorporating human demonstrations, feedback, mechanisms, and knowledge into autonomous-driving systems~\cite{Li2025}. One potential solution to enhance the safety of \acp{AV} is integrating human risk-perception and decision-making capabilities with existing algorithms using \ac{BCI}. Rather than acting directly as drivers, humans assist the computer by providing cognitive supervision signals~\cite{belcher2022eeg}.

\subsection{Related Work}

\subsubsection{Neurophysiological studies of driving risk cognition}
Multiple neuroimaging techniques, including \ac{MEG}~\cite{Pizzo2021Author,baillet2017magnetoencephalography,hamalainen1993magnetoencephalography}, \ac{EEG}~\cite{di2019eeg,lee2020analysis,morando2018reference,zhao2022research,huang2021deep,zeeb2016take,li2023drivers,Brosnan2020-ed}, \ac{fMRI}~\cite{crosson2010functional,dimou2013systematic,Holton2024-xx,Yoo2022-mc}, and \ac{fNIRS}~\cite{pinti2020present,geissler2023functional,perello2023drivers} have been used to study brain activity and cognitive states. These modalities provide complementary advantages. \ac{MEG} and \ac{fMRI} offer rich spatial or source-level information, while \ac{fNIRS} has been widely used to measure prefrontal activity related to workload, distraction, perceived risk, and traffic complexity~\cite{baker2021evaluation,foy2018mental,6849496,perello2023drivers,Zhang2023fNIRS,wang2023driving}. Borowsky and Oron-Gilad further distinguished risk perception, i.e., the subjective evaluation of handling ability, from hazard perception, i.e., real-time danger detection, and suggested a sequential relationship between them~\cite{borowsky2013exploring}. These studies provide important neurocognitive evidence that traffic risk is reflected in human brain activity. However, many of them focus on explaining cognitive responses rather than developing an \ac{EEG} decoding framework that can jointly support early risk prediction and immediate danger identification.

\subsubsection{EEG-based driving safety and hazard-related decoding}
Compared with \ac{MEG}, \ac{fMRI}, and \ac{fNIRS}, \ac{EEG} provides high temporal resolution, portability, and potential real-time usability, making it suitable for decoding fast-evolving cognitive responses in driving scenarios~\cite{Pizzo2021Author,di2019eeg,crosson2010functional,pinti2020present}. \ac{EEG} has been applied to driving behavior recognition~\cite{YANG201830,ZHAO2022106665,MA2024107769}, emergency braking intention detection~\cite{nguyen2019detection,8026574}, autonomous decision-making enhancement~\cite{Shin2022wearable}, driver fatigue monitoring~\cite{9606579,10904061}, and shared vehicle control~\cite{8026574,lu2019model}. In parallel, recent human-centered vehicle-control studies have incorporated
human-emulated behavior and driver intention into automated or shared-control strategies to improve safety and human--machine coordination
~\cite{Liu2026,Xie2026}. Previous work has also shown that \ac{EEG} exhibits stronger low-frequency activity below 10\,Hz during dangerous events than during normal driving~\cite{zhang2023EEG}. These findings indicate that \ac{EEG} contains information related to perceived traffic risk, hazard processing, and behavioral readiness. Nevertheless, most existing EEG-based driving studies are driver-centered and are closely coupled with driving control, braking intention, fatigue, or takeover behavior. In highly automated vehicles, the human role gradually shifts from active driver to passive passenger, whose cognition is more observational and anticipatory. This passenger-centered setting remains underexplored.

\subsubsection{EEG decoding models}
From the modeling perspective, early \ac{EEG} decoding studies commonly relied on handcrafted spatial-spectral features and conventional classifiers, such as \ac{CSP} with \ac{LDA} or \ac{SVM}. Recent studies have increasingly adopted deep learning models for \ac{EEG} classification~\cite{SHARMA2022103101,EEGDLRE,DeepLearningEEG}. \ac{CNN}-based models, including ShallowConvNet, DeepConvNet~\cite{DeepLearningEEG}, and EEGNet~\cite{EEGNet}, learn spatial-temporal representations directly from \ac{EEG} and have become widely used baselines in \ac{BCI} research. Multi-scale \ac{CNN} and transfer-learning designs have further improved \ac{EEG} feature extraction and cross-subject robustness~\cite{ROY2022103496,ROY2022105347}. In addition to \acp{CNN}, recurrent models such as \ac{LSTM} and \ac{GRU} have been introduced to capture temporal dependencies in \ac{EEG} sequences~\cite{EEGDLRE,EEGTime}, while hybrid feature-driven methods such as XGB-DIM combine global and local spatial-temporal filters to extract discriminative \ac{EEG} representations~\cite{bib8}. More recent temporal models, including Transformer-based and ConvLSTM-based \ac{EEG} decoders, further emphasize long-range dependency modeling and spatiotemporal sequence representation.

These prior models provide strong foundations for \ac{EEG} decoding, but they are usually designed for generic \ac{BCI} tasks, motor imagery, \ac{RSVP}, or driver-state recognition. They do not explicitly encode the traffic-risk evolution from potential risk to instantiated danger, nor do they jointly model pre-onset risk anticipation and near-onset danger identification in a passenger-centered \ac{AV} setting. Therefore, in this study, we compare our method with representative CNN-based, feature-driven, Transformer-based, and ConvLSTM-based \ac{EEG} baselines to evaluate whether the proposed framework provides advantages beyond standard \ac{EEG} decoding architectures.

\subsubsection{Positioning of the proposed framework}
In summary, existing studies have established that human neural activity reflects driving-related workload, risk perception, and hazard processing. However, three gaps remain. First, most driving neuroergonomics studies focus on drivers~\cite{perello2023drivers,nguyen2019detection,wang2023driving,8026574,9606579,10904061,10706793}, whereas passengers in highly automated vehicles have different roles, lower control responsibility, and more observational cognitive responses~\cite{Merat2019Out,Endsley2016From}. Second, most EEG-based methods focus on immediate hazard detection after a specific event emerges~\cite{nguyen2019detection,Shin2022wearable}, while early risk prediction before hazardous behavior is less studied. Third, existing \ac{EEG} decoders usually treat classification as a single isolated task, rather than modeling the sequential relationship between risk anticipation and danger recognition.

To address these gaps, our work proposes a passenger-centered \ac{EEG} decoding framework that differs from prior studies in three aspects. First, the \ac{PCM} links traffic stages with passenger neural responses, providing a neurocognitive basis for task formulation. Second, \ac{RSL} explicitly models the transition from Safe to Low-Risk and High-Risk conditions, thereby coupling \ac{RP} and \ac{DI}. Third, the \ac{PEDS} implements this formulation using an electrode-informed \ac{3D-CRNN} architecture that captures spatial topology and temporal dynamics for both pre-onset \ac{RP} and near-onset \ac{DI}. This design positions the proposed framework not only as another \ac{EEG} classifier but as a task-aware decoding framework for passenger hazard perception in highly automated vehicles.

\subsection{Contributions}
To address these gaps, we propose a passenger-centered \ac{EEG} decoding framework for highly automated vehicles, with the following four levels of novelty: 

\begin{itemize}
    \item \textbf{Data novelty}: we present a 45-hour passenger-centered \ac{EEG} dataset from 15 participants, comprising 14 distinct traffic scenarios that cover the most common traffic conditions. 
    \item \textbf{Neurocognition novelty}: we introduce the \ac{PCM} to explain how passenger brain activity evolves from calm monitoring to risk anticipation and danger recognition, with respect to traffic conditions. 
    \item \textbf{Task-design novelty}: we propose \ac{RSL} to explicitly encode the progression from safe to low-risk and high-risk traffic conditions, thereby coupling \ac{RP} and \ac{DI} in a sequential manner. 
    \item \textbf{Model novelty}: we develop the \ac{PEDS} based on an electrode-informed \ac{3D-CRNN}, which jointly captures neural evidence for pre-onset risk prediction and post-onset danger identification.
\end{itemize}

Quantitatively, our framework also delivers clear performance advantages. In Risk Prediction, the \ac{RP} branch of the proposed \ac{3D-CRNN} achieves a \ac{BA} of 95.3\%, demonstrating that passenger \ac{EEG} contains robust pre-event information for anticipatory risk decoding. In single-subject \ac{DI}, \ac{3D-CRNN} with \ac{RSL} achieves the best overall result, reaching 85.0\% \ac{BA}, outperforming the compared baselines, including EEGNet (80.3\%), ShallowConvNet (80.3\%), DeepConvNet (77.6\%), XGB-DIM (73.4\%), CTNet (68.2\%), and DSC-ConvLSTM (72.2\%). Event-wise analysis further shows that \ac{3D-CRNN} with \ac{RSL} achieves the best \ac{DI} performance across \ac{AEB}, cut-in, and pedestrian scenarios, with \ac{BA} values of 77.8\%, 88.0\%, and 79.9\%, respectively. In addition, the benefit of the proposed sequential formulation is supported by the ablation study, where incorporating \ac{RSL} improves \ac{BA} for \ac{3D-CRNN} and most baselines in single-subject \ac{DI}. In terms of generalizability, it reaches a \ac{BA} of 77.0\% in the cross-session setting, 77.4\% on seen subjects, and 64.9\% on unseen subjects, achieving the highest Precision, Recall, and F1-score in cross-subject evaluation but the second-highest \ac{BA} on unseen subjects. This shows that our model has strong robustness toward both intra-subject and inter-subject variability.

\subsection{Paper Organization}
The rest of this paper is organized as follows. In Section~\ref{sec:Method}, we describe the experimental paradigm, data analysis, and methodology. Section~\ref{sec:Results} presents the results, followed by a discussion of our findings, challenges, and future research directions in Section~\ref{sec:Discussion}. Finally, Section~\ref{sec:Conclusion} summarizes our contributions and provides closing remarks.
The PassengerEEG dataset~\cite{PassengerEEG2025} is hosted on IEEE DataPort\footnote{\url{https://ieee-dataport.org/documents/passengereeg-eeg-dataset-passenger-hazard-perception-avs}}, and the related code is available on GitHub\footnote{\url{https://github.com/SOTIF-AVLab/EEG2023}}.
\section{Methodology}
\label{sec:Method}

\begin{figure}[tbp]
\centering
\includegraphics[trim=16 0 440 0, clip, width=\textwidth]{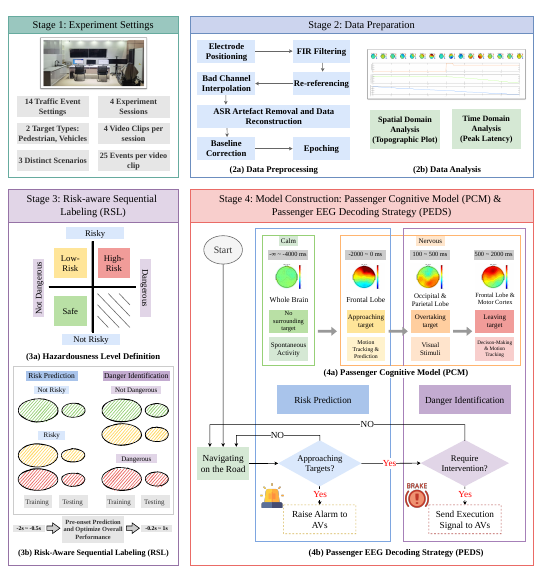}
\caption{Overall workflow of the proposed EEG-driven decoding framework. Stage 1 summarizes the simulator experimental settings and the dataset. Stage 2 shows the \ac{EEG} preprocessing and analysis of cognitive insights. Stage 3 presents \ac{RSL}: hazardousness is defined on Risky/Not Risky and Dangerous/Not Dangerous axes to form three valid levels (Safe, Low-Risk, High-Risk) and corresponding labels for \ac{RP} and \ac{DI}. Stage 4 illustrates model construction, including the \ac{PCM} and \ac{PEDS}, and the resulting \ac{RP} and \ac{DI}. Dashed boxes indicate the decoded auxiliary signal for downstream \ac{AV} actions, i.e., alarms and executions, shown for context but not studied in this work.}
\label{fig1}
\end{figure}

\subsection{Problem Formulation}
Fig.~\ref{fig1} outlines the overall research flow, including Experiment Settings, Data Analysis, Data Preparation, and Model Construction. Our goal is to leverage passenger \ac{EEG} as an auxiliary supervision signal to enhance \ac{AV} safety development by decoding two temporally distinct cognitive processes:
\begin{itemize}
    \item \textbf{\emph{Hazard anticipation}} before an event manifests.
    \item \textbf{\emph{Hazard recognition}} after the event is instantiated.
\end{itemize}

Accordingly, we address: \\
\textbf{Q1:} What cognitive insights can be derived from passenger \ac{EEG} during evolving traffic interactions? \\
\textbf{Q2:} How should we formulate decoding tasks to exploit these insights for AV safety?

\subsubsection{Operational Event Onset}
Let $t_0$ denote the event onset, defined as the earliest time at which the target may instantiate a hazardous maneuver (e.g., cut-in, sudden crossing, emergency-braking conflict) that would require a prompt ego-vehicle response (e.g., deceleration or evasive action) to maintain safety.

\subsubsection{Input Data}
For each event-aligned sample, the model input is a multichannel \ac{EEG} segment
\[
X \in \mathbb{R}^{C \times T},
\]
where $C$ is the number of \ac{EEG} channels and $T$ is the number of time samples in the analysis window. We align the \ac{EEG} to the event onset time $t_0$ and re-reference the time to $t_0=0$. Hence, $X(:,t)$ denotes the \ac{EEG} observation at relative time $t$.

\subsubsection{Hazardousness Level Definition}

As illustrated in Fig.~\ref{fig1} (3a), we model the hazardousness of an ego--target interaction using two binary axes: (i) \emph{potential hazard} (Risky vs.\ Not Risky) and (ii) \emph{instantiated hazard} (Dangerous vs.\ Not Dangerous).
The three risk levels (\textit{Safe}, \textit{Low-Risk}, \textit{High-Risk}) correspond to the valid quadrants in Fig.~\ref{fig1} (3a).

\paragraph{Y-Axis: Risky vs.\ Not Risky (potential hazard; pre-onset)}
Let $\mathrm{TTE}(t)$ denote the time-to-event to $t_0$ at time $t<0$ and let $\tau=5\,\mathrm{s}$.
We define the indicator as follows:
\begin{definition}
    \label{def:risky}
    \textbf{Risky} -- \textbf{\textit{pre-onset potential hazard state}}: target--ego interaction context in which a nearby target may become hazardous, but no immediate hazardous maneuver has yet been instantiated.
    Mathematically, 

    \begin{equation}
    r(t) \triangleq \mathbb{I}\!\left(\mathrm{TTE}(t) < \tau\right), \quad t<0,
    \label{eq:risky_indicator}
    \end{equation}
\end{definition}

Otherwise, it is Not Risky.

\paragraph{X-Axis: Dangerous vs.\ Not Dangerous (instantiated hazard; onset/post-onset)}
We define a hazardous-behavior existence indicator $h(t)$, where $h(t)=1$ if a hazardous maneuver is present/active at time $t$ (e.g., cut-in, sudden crossing, emergency-braking conflict), and $h(t)=0$ otherwise.
We define the indicator as
\begin{definition}
    \label{def:dangerous}
    \textbf{Dangerous} -- \textbf{\textit{post-onset immediate hazard state}}: the target vehicle has instantiated a hazardous maneuver that poses an immediate threat if the ego vehicle does not respond.
    Mathematically,
    \begin{equation}
    d(t) \triangleq \mathbb{I}\!\left(t \ge 0 \;\wedge\; h(t)=1\right),
    \label{eq:dangerous_indicator}
    \end{equation}
\end{definition}

Otherwise, it is Not Dangerous.

Notably, these two situations occur sequentially: the ego vehicle first approaches other traffic participants (entering a \textbf{Risky} state) before potentially interacting with them in a hazardous manner (transitioning to a \textbf{Dangerous} state). 

\subsubsection{Risk-aware Sequential Labeling (RSL)}
As illustrated in Fig.~\ref{fig1} (3a), we describe each ego--target interaction using two binary indicators with a sequential constraint: 
\textbf{Risky vs.\ Not Risky} represents the presence of a potential hazard before event onset, whereas \textbf{Dangerous vs.\ Not Dangerous} represents whether the potential hazard is instantiated after event onset. 
\ac{RSL} converts this constrained two-indicator representation into three valid hazardousness levels, corresponding to the three unhatched quadrants in Fig.~\ref{fig1} (3a):
\begin{equation}
\ell \triangleq 
\begin{cases}
0, & (r,d)=(0,0) \quad \text{Safe}\\
1, & (r,d)=(1,0) \quad \text{Low-Risk}\\
2, & (r,d)=(1,1) \quad \text{High-Risk}~,
\end{cases}
\label{eq:risk_level_mapping}
\end{equation}
where $(r,d)=(0,1)$, namely, Not Risky but Dangerous, is excluded by construction as the hatched region in Fig.~\ref{fig1} (3a).
In this taxonomy, Dangerous is not defined as an independent category from Risky; instead, it is treated as a stricter state nested within Risky. 
Formally, $d=1$ implies $r=1$, and therefore all Dangerous samples are included in the Risk class for \ac{RP}. 
Consequently, the \ac{RP} task separates Safe samples from both Low-Risk and High-Risk samples, whereas the \ac{DI} task separates High-Risk samples from Safe and Low-Risk samples. 
This design enforces the sequential logic that an interaction should first enter a potential hazard stage, i.e., Risky, before an immediate hazard can be instantiated, i.e., Dangerous. 
Fig.~\ref{fig1} (3b) illustrates how this sequential structure is used to form coherent labels for the two tasks.

\subsubsection{Decoding Tasks}
Based on the definition in Fig.~\ref{fig1} (3a) and the labeling strategy in Fig.~\ref{fig1} (3b), we formulate two binary decoding tasks that correspond to distinct temporal stages of hazard evolution and match the two outputs in Fig.~\ref{fig1} (4b).

\begin{definition}
\textbf{Risk Prediction} (\ac{RP}): Decodes whether the current traffic interaction is likely to evolve into a risk-relevant situation during the \emph{pre-event} approach phase (\emph{Risky} stage). 
\end{definition}
\begin{equation}
    y_{\mathcal{RP}} \in \{0,1\}, \qquad 
    y_{\mathcal{RP}} = \mathbb{I}(\ell \ge 1),
    \label{eq:y_rp}
\end{equation}
i.e., \ac{RP} is positive for both Low-Risk and High-Risk quadrants in Fig.~\ref{fig1} (3a).

\begin{definition}
\textbf{Danger Identification} (\ac{DI}): Identifies whether an \emph{immediate hazard} has been instantiated (Dangerous stage) around onset/post-onset, supporting real-time hazard assessment.
\end{definition}
\begin{equation}
    y_{\mathcal{DI}} \in \{0,1\}, \qquad 
    y_{\mathcal{DI}} = \mathbb{I}(\ell = 2),
    \label{eq:y_di}
\end{equation}
i.e., \ac{DI} is positive only for the High-Risk quadrant (Dangerous) in Fig.~\ref{fig1} (3a).

\noindent\textbf{Sequential constraint.} Since Dangerous implies Risky in Fig.~\ref{fig1} (3a), the labels satisfy $y_{\mathcal{DI}}=1 \Rightarrow y_{\mathcal{RP}}=1$ by construction.

\subsubsection{Objective}
Given an event-aligned \ac{EEG} segment $X \in \mathbb{R}^{C\times T}$, our objective is to learn a decoding function
\begin{equation}
f_{\theta}: X \mapsto \left(\hat{y}_{\mathcal{RP}},\,\hat{y}_{\mathcal{DI}}\right),
\label{eq:objective_mapping}
\end{equation}
that jointly predicts the risk prediction outcome ($\mathcal{RP}$) and the danger identification outcome ($\mathcal{DI}$) defined in~\eqref{eq:y_rp}--\eqref{eq:y_di}.
Consistent with Fig.~\ref{fig:time_scheme}, the two outputs are produced from task-specific \ac{EEG} windows:
\begin{equation}
X_{\mathcal{RP}} = X\big|_{t\in \mathcal{W}_{\mathcal{RP}}}, \quad
X_{\mathcal{DI}} = X\big|_{t\in \mathcal{W}_{\mathcal{DI}}},
\label{eq:task_windows}
\end{equation}
where $\mathcal{W}_{\mathcal{RP}}=[-2.0,-0.5]$~s is strictly pre-onset to ensure causal \ac{RP} decoding, and $\mathcal{W}_{\mathcal{DI}}=[-0.2,1.0]$~s captures onset/immediate post-onset activity for $\mathcal{DI}$. The selection of the time windows is discussed in \S~\ref{sec:time-window-selection}.

The proposed framework comprises three components, as represented in Fig.~\ref{fig1}:
\begin{itemize}
    \item \textbf{\acf{RSL}}, which maps the two-axis interaction state $(r,d)$ to a valid risk level $\ell$ and task labels $(y_{\mathcal{RP}},y_{\mathcal{DI}})$, as shown in Fig.~\ref{fig1} (3).
    \item \textbf{\acf{PCM}}, which links traffic stages to expected neural dynamics and motivates the task design, as shown in Fig.~\ref{fig1} (4a).
    \item \textbf{\acf{PEDS}}, which implements $f_\theta$ with a joint \ac{3D-CRNN} decoder to produce simultaneous \ac{RP}/\ac{DI} predictions that can be integrated into an AV safety pipeline, as shown in Fig.~\ref{fig1} (4b).
\end{itemize}

\subsection{Data Collection Setup}
\begin{figure}[tbp]
    \centering
    \includegraphics[trim=0 40 0 10, clip, width=\textwidth]{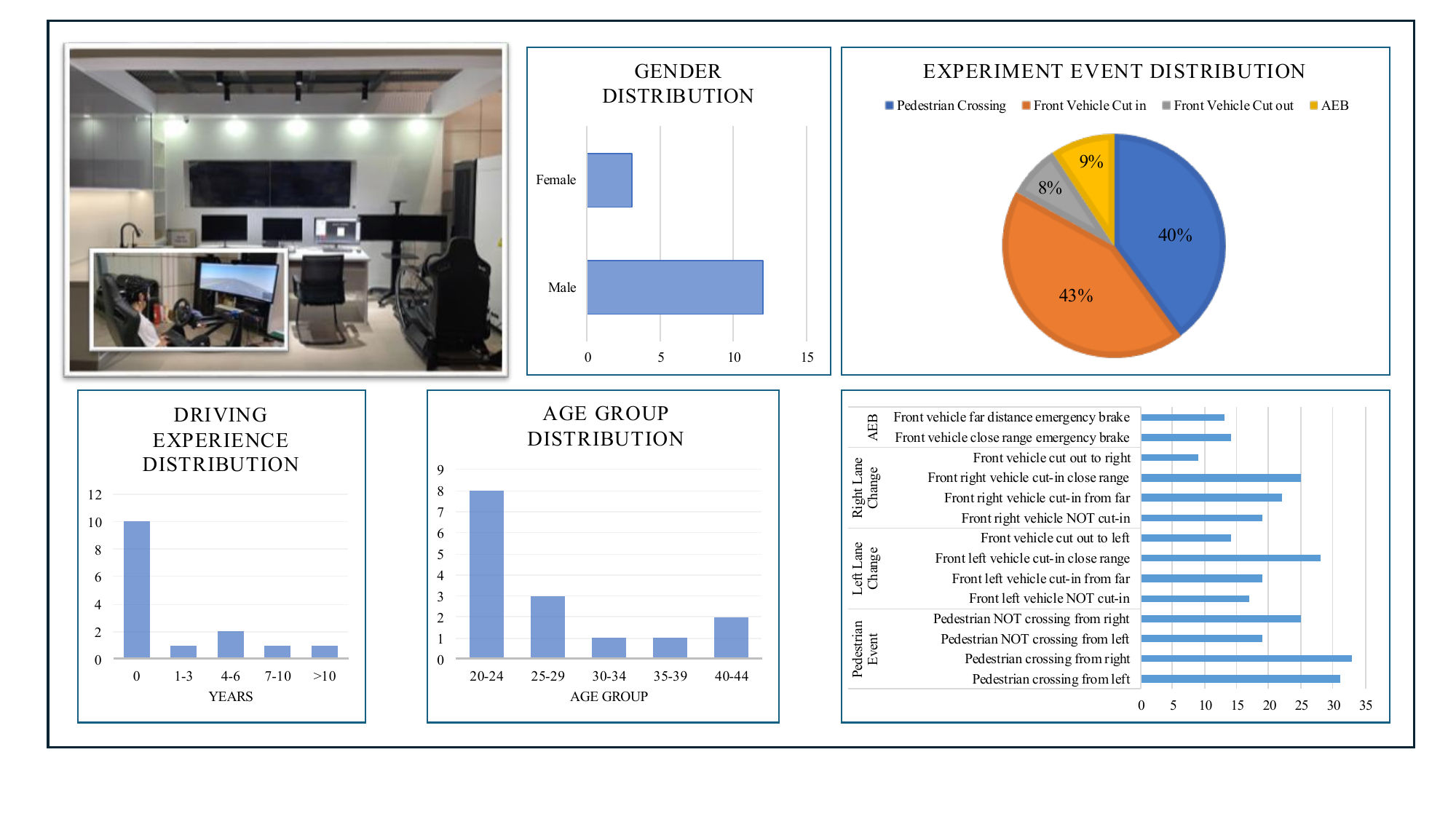}
    \caption{Overview of the experimental setting and dataset composition. The figure includes the driving simulator and \ac{EEG} acquisition environment, participant demographic distributions (gender, age group, and driving experience), and the composition of the simulated traffic events used in the experiment.}
    \label{fig:experiment_overview}
\end{figure}

Fig.~\ref{fig:experiment_overview} provides an overview of the data collection setup, including the simulator and \ac{EEG} acquisition environment, participant demographics, and the distribution of traffic events used in the study.

\subsubsection{Ethics Statement}
This study complied with the Declaration of Helsinki and was approved by the Institutional Review Board of Tsinghua University, China. Participants were informed of their right to ask questions about the study and were assured that their personal and identifiable data would remain confidential. Additionally, they were encouraged to take breaks whenever necessary and were free to end the session at any time for any reason. All individuals provided written informed consent after receiving a full explanation of the study procedures. Participants were also compensated for their participation.

\subsubsection{Participants}
The participants in this study were recruited from the general population. In total, 15 participants were enrolled, including 12 males and 3 females, with a mean age of 27.8 years (standard deviation: 6.625) and an age range of 22 to 42 years. The demographic distributions of gender, age group, and driving experience are summarized in Fig.~\ref{fig:experiment_overview}. Among the 15 recruited participants, data from 11 participants were deemed valid for analysis.

None of the participants reported any disease or susceptibility to simulator sickness. Since all participants were treated as passengers rather than drivers in this study, driving experience was not a prerequisite for participation. Nevertheless, to increase the diversity of the dataset, participants with different levels of driving experience were included. Specifically, 6 participants were experienced drivers (5 males and 1 female) with at least 1 year of driving experience and no reported history of accidents or traffic violations.

\subsubsection{Data Collection Sessions}
The experiment was conducted in our driving simulation laboratory in a controlled, quiet, and distraction-free environment. Participants were seated in a driving simulator equipped with a dashboard display that presented first-person traffic video clips, while \ac{EEG} signals were recorded throughout the experiment. The physical setup of the simulator and \ac{EEG} acquisition environment is shown in Fig.~\ref{fig:experiment_overview}.

The traffic video clips were generated using the \ac{VTD} platform (Hexagon AB, Stockholm, Sweden) to simulate diverse traffic scenarios. As summarized in Fig.~\ref{fig:experiment_overview}, the dataset included multiple event categories, such as pedestrian crossing, front-vehicle cut-in, front-vehicle cut-out, and emergency braking. During the experiment, participants were instructed to press the space bar upon hearing a designated auditory cue, which served to maintain their attention throughout the session and to mark moments when they perceived a dangerous situation. No additional responses or actions were required.

The experiment consisted of four separate sessions conducted on different days. To complete the experiment, each subject was required to attend all four sessions, watching a total of 16 unique video clips, with each session consisting of 4 clips. To minimize fatigue, participants were limited to one session per day and allowed to take a break after each video clip.

\subsubsection{Scenario Specifications}
\begin{figure}[H]
\centering
\begin{minipage}[t]{0.48\textwidth}
    \vspace{0pt}
    \centering
    \captionsetup{font=footnotesize,skip=4pt}
    \scriptsize
    \setlength{\tabcolsep}{0.012\linewidth}
    \renewcommand{\arraystretch}{1.04}
    \begin{tabular}{@{}
        >{\centering\arraybackslash}p{0.06\linewidth}
        >{\raggedright\arraybackslash}p{0.17\linewidth}
        >{\raggedright\arraybackslash}p{0.32\linewidth}
        >{\centering\arraybackslash}p{0.10\linewidth}
        >{\centering\arraybackslash}p{0.13\linewidth}
        >{\centering\arraybackslash}p{0.10\linewidth}
    @{}}
    \toprule
    \textbf{ID} & \textbf{Type} & \textbf{Action} & \textbf{Dir.$^{\dagger}$} & \textbf{Dist.} & \textbf{Risk} \\
    \midrule
    1  & Pedestrian & Cross  & L  & N/A   & High   \\
    2  & Pedestrian & Cross  & R  & N/A   & High   \\
    3  & Pedestrian & Does Not Cross  & L  & N/A   & Low    \\
    4  & Pedestrian & Does Not Cross  & R  & N/A   & Low    \\
    5  & Vehicle    & No Cut-in & L  & N/A   & Low    \\
    6  & Vehicle    & No Cut-in & R  & N/A   & Low    \\
    7  & Vehicle    & Cut-in & L  & Close & High   \\
    8  & Vehicle    & Cut-in & R  & Close & High   \\
    9  & Vehicle    & Cut-in & L  & Far   & High   \\
    10 & Vehicle    & Cut-in & R  & Far   & High   \\
    11 & Vehicle    & Cut-out without AEB$^{\ddagger}$  & F  & Close & Low    \\
    12 & Vehicle    & Cut-out without AEB  & F  & Far   & Low    \\
    13 & Vehicle    & Cut-out with AEB  & F  & Close & High   \\
    14 & Vehicle    & Cut-out with AEB  & F  & Far   & High   \\
    \bottomrule
    \end{tabular}
    \captionof{table}{Details of traffic events, including ID, Target Type, Action, Direction (Dir.), Distance (Dist.), and Risk Level.}
    \label{tab:events_table}
    \vspace{2pt}
    {\scriptsize\raggedright $^{\dagger}$L: Left, R: Right, F: Front; $^{\ddagger}$AEB: Automated Emergency Braking.\par}
\end{minipage}\hfill
\begin{minipage}[t]{0.48\textwidth}
    \vspace{0pt}
    \centering
    \captionsetup{font=footnotesize,skip=4pt}
    \includegraphics[width=\linewidth,trim={0cm 0cm 10cm 0cm},clip]{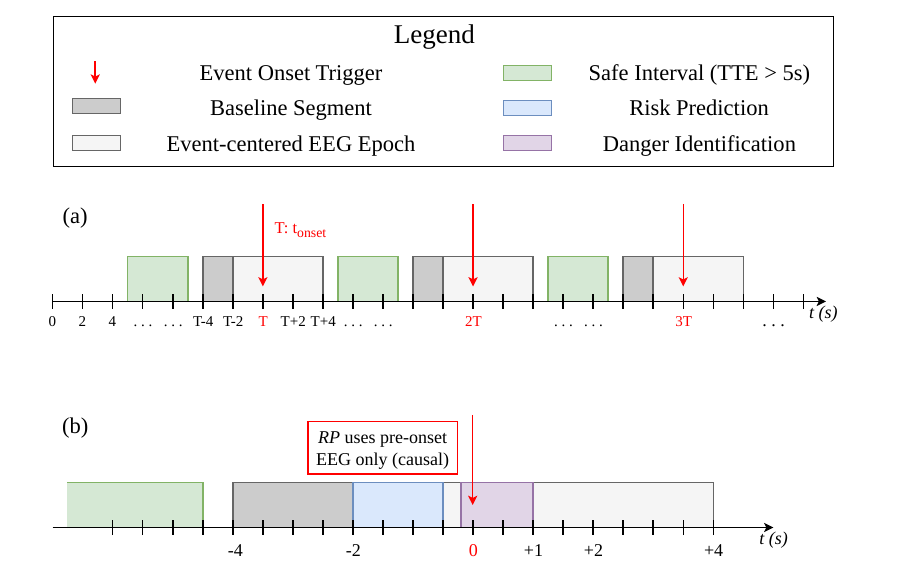}
    \captionof{figure}{Event-centered \ac{EEG} segmentation (top) and task-specific windows for \ac{RP} and \ac{DI} (bottom).}
    \label{fig:time_scheme}
\end{minipage}
\end{figure}

The events created in the experiment span the most common traffic scenarios, including pedestrian crossing, vehicles cutting in, and leading vehicles cutting out with or without \ac{AEB}. Detailed information about the experimental setting is shown in Table~\ref{tab:events_table}. Additional specifications, such as vehicle speed and ego--target interaction distance, are presented in Appendix~\ref{app:scenario-specifications}.

In Fig.~\ref{fig:time_scheme}(a), long driving clips contain repeated hazard events with onsets at $T, 2T, 3T, \dots$. For each onset, an event-centered \ac{EEG} epoch $[T-4,\,T+4]$ is extracted, including a baseline segment $[T-4,\,T-2]$. Additional safe samples are collected from intervals with $\mathrm{TTE} > 5$\,s. Fig.~\ref{fig:time_scheme}(b) presents the local timeline for a single event, re-referenced such that hazard onset is at $t=0$. \ac{RP} uses only the pre-onset \ac{EEG} window $[-2.0,\,-0.5]$ s (causal setting), while \ac{DI} uses the onset/immediate post-onset window $[-0.2,\,+1.0]$ s. This design enforces causal prediction for \ac{RP} and avoids post-onset \ac{EEG} leakage.

Each video clip lasts approximately 15\,min and includes 25 event-specific traffic scenarios, interspersed with event-free segments, randomly selected from 14 event types. By the end of the experiment, we expect to collect 400 event-related \ac{EEG} segments and approximately 4\,h of \ac{EEG} data per participant. The timeline of a typical video clip and event setup is illustrated in Fig.~\ref{fig:time_scheme}.

In this experiment, $T$ denotes the event onset moment, such as a pedestrian on the sidewalk suddenly rushing across the road. To introduce variability, the time interval between consecutive events ranges from 40 to 60\,s.

Each event-specific scenario contributes an 8-s data segment, highlighted in gray. Within each segment, the first 2\,s (dark gray) serve as a baseline to prevent signal drift, while the remaining 6\,s (light gray) capture event-specific neural activity, including potential interactions with other traffic participants. Additionally, data segments are color-coded based on their relevance to classification tasks: blue represents pre-event data, typically used for Risk Prediction, while purple marks the moment of event onset, used for \ac{DI}. Green segments correspond to Safe intervals between events, defined as moments where $\mathrm{TTE}$ $>$ 5\,s.

\subsubsection{Data Preprocessing}
\ac{EEG} data are collected using a 64-channel EEG Quik-Cap, a product of NeuroScan, designed with soft, flexible material to conform to each participant's head for optimal electrode contact. The electrodes, made of sintered Ag/AgCl, provide high conductivity, ensuring high-quality signal acquisition. We adjust the conductive paste to ensure that the resistance of all electrodes before each experimental session is lower than $10\,\mathrm{k}\Omega$. Electrodes are positioned according to the International 10--20 System, covering the frontal, temporal, occipital, and lateral regions of the scalp to capture comprehensive brain activity.

Real-time data acquisition is performed using NeuroScan\textsuperscript{TM} Curry 8 software (Compumedics\textsuperscript{\textregistered}, Abbotsford, Victoria, Australia) at a sampling rate of 1000\,Hz. The \ac{EEG} signals are amplified using a SynAmps RT 64-channel \ac{EEG} amplifier, which enables high-fidelity recording of low-amplitude brain signals.

MATLAB (MathWorks\textsuperscript{\textregistered}, Natick, MA, USA) and its EEGLAB~\cite{delorme2004eeglab} functions are widely utilized tools for the processing and analysis of \ac{EEG} signals. To enhance the signal quality by eliminating drift, high-frequency distortion, and power line noise, an \ac{FIR} bandpass filter from 0.5\,Hz to 50\,Hz is applied to the \ac{EEG} signals. After filtering, the \ac{EEG} signals are re-referenced to the average of the left mastoid (M1) and right mastoid (M2) electrodes, typically placed behind the ears over the mastoid processes, to further reduce noise and increase clarity. Damaged and abnormal electrodes are automatically removed and interpolated to maintain signal integrity. Subsequently, \ac{ASR} is used to suppress eye blinking and other muscle artifacts. In our application, ASR offers advantages over \ac{ICA}, as it is notably more time-efficient, enabling online and real-time application while reconstructing signals free of unwanted artifacts~\cite{Chang_Hsu_Pion-Tonachini_Jung_2018}. During the data segmentation process, all event-related signals are extracted based on the event start time in the traffic video clip. Each segment encompasses a period from 4\,s before the event onset to 4\,s afterward. Other non-event-specific data are randomly sampled from the Safe interval between consecutive events. Finally, we perform baseline correction to prevent potential signal drift.

\subsection{Passenger Cognitive Model (PCM)}
In this section, we propose \ac{PCM} to answer the first question---\textit{What cognitive insights can be derived from passenger \ac{EEG} signals?} \S~\ref{sec:pcm-topographic-analysis} describes the topographic-map analysis used to identify brain-activity patterns under different traffic conditions. \S~\ref{sec:pcm-traffic-neural-connection} relates these neural patterns to the evolution of traffic events and their associated risk levels. Finally, \S~\ref{sec:pcm-model} consolidates these observations into the \ac{PCM}, which links traffic scenarios, mental states, neural responses, cognitive processes, and the corresponding decoding tasks.

\subsubsection{\textbf{Brain Activity Analysis -- Topographic Maps}}
\label{sec:pcm-topographic-analysis}
The topographic map is an intuitive method for visualizing active brain regions over a selected time window. We use EEGLAB plugins in MATLAB, a widely used tool for generating topographic maps~\cite{delorme2004eeglab}. Given that \ac{EEG} signals are inherently weak and highly intolerant to noise and artifacts, the low \ac{PSNR} of a single-trial \ac{EEG} recording makes direct one-to-one comparisons of topographic plots between different traffic events unreliable and potentially uninformative. Hence, we obtain clearer and more meaningful results by taking the average of specific data groups, as event-related patterns are superimposed and noise is canceled out.

To comprehensively analyze the passengers' risk-aware cognitive process, we employ two distinct comparison strategies. First, by comparing \ac{EEG} segments from event-free periods (i.e., Safe periods) with event-related \ac{EEG} data, we can infer passengers' perception and awareness of potential risks. Second, to interpret passengers' danger-identification cognition while controlling for other variables, we compare \ac{EEG} segments corresponding to specific events of interest with data recorded in similar traffic scenarios but at different risk levels. For instance, \ac{EEG} segments from Event 7 (a High-Risk cut-in scenario) are compared with those from Event 5 (a Low-Risk cut-in scenario) to examine brain activity associated with observing specific hazardous behaviors.

Since the time-window selection influences topographic maps, we adopt two different temporal resolutions: 50\,ms, which captures a more detailed temporal evolution of brain activity, and 400\,ms, which highlights general activation patterns over a broader time scale. The potential range is confined to the interval \([-6, 6] \,\mu\mathrm{V}\) to ensure that the visualization remains consistent and allows for clear interpretation across different subjects, events, and risk levels.

\subsubsection{\textbf{Establish Connection}}
\label{sec:pcm-traffic-neural-connection}
\begin{figure}[tbp]
    \centering
    \includegraphics[trim=0 160 300 0, clip, width=\textwidth]{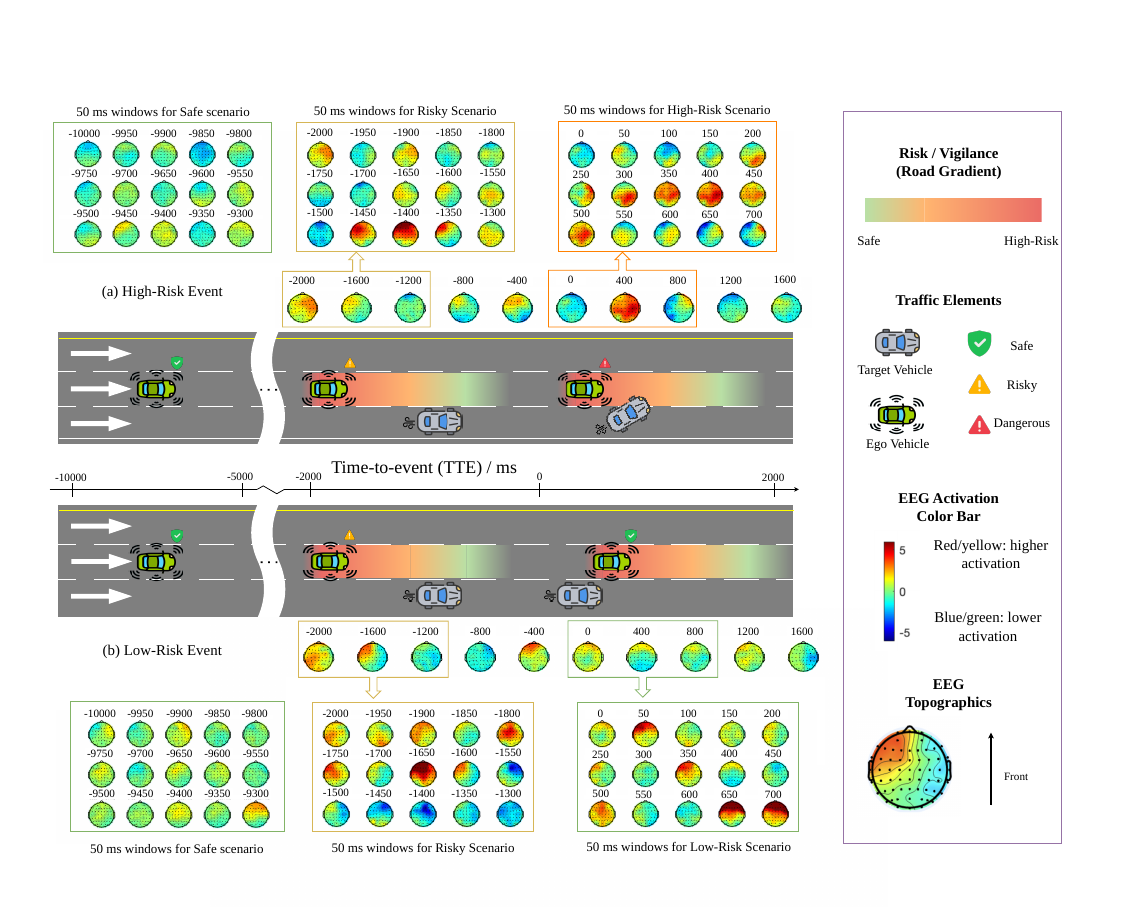}
    \caption{Visualization of \ac{EEG} topographic maps aligned with traffic scenarios over a unified timeline. The middle section illustrates two driving scenarios: a high-risk cut-in event (Event 8; top) and a low-risk cut-in event (Event 6; bottom). The passenger vigilance zone is color-coded, with green indicating safety and yellow-to-red representing increasing perceived risk. The surrounding \ac{EEG} topographic maps depict neural activity at different time points, using two temporal resolutions: 50\,ms (detailed neural dynamics) and 400\,ms (general activation patterns). Warmer colors (red) indicate increased brain activity, and cooler colors (blue) denote lower neural activation.}
    \label{fig:topo_scene}
\end{figure}

Fig.~\ref{fig:topo_scene} visually aligns \ac{EEG} topographic maps with traffic scenarios on a unified timeline, from 10\,s preceding the event onset to 2\,s afterward, helping to establish the connection between neural responses and traffic conditions. In this figure, we compare Event 6 (a Low-Risk cut-in scenario) with Event 8 (a High-Risk cut-in scenario) to demonstrate how cognitive load and risk awareness fluctuated based on the nature of traffic events.

The event onset is defined as the moment when a target vehicle initiates a hazardous maneuver that may threaten the ego vehicle’s driving safety. For instance, a vehicle in the adjacent lane suddenly cuts into the ego lane at a short distance. When no visible or distant target vehicles are present ahead ($\mathrm{TTE}$ $>$ 5\,s), this period is classified as Safe, where \ac{EEG} activity remains stable, indicating a calm cognitive state.

As the ego vehicle approaches a front vehicle, heightened neural activity emerges. Between 1200\,ms and 1600\,ms before event onset, a significant positive fluctuation in the frontal lobe is observed, suggesting increased passenger attention to potential risks. This trend is evident in both High-Risk and Low-Risk scenarios, appearing at 1300--1450\,ms and 1550--1900\,ms, respectively.

In Event 8, when the target vehicle initiates a lane change (cut-in), the \ac{EEG} signals reveal a strong positive fluctuation in the occipital and parietal lobes, with a latency of approximately 200--300\,ms. This response aligns with the P300 component, indicating that the passenger recognizes an unexpected and potentially hazardous action.

In contrast, in Event 6, a non-cut-in scenario, only minor frontal lobe activity is observed, and no significant P300 component is present. This suggests the absence of a sudden stimulus, as the passenger does not experience an unexpected change in the traffic environment.

We visualize the passenger's nervousness as a color map in the figure, representing their level of alertness based on the distance between the ego vehicle and the target vehicles. In this map, green indicates a safe and relaxed state, while colors transition to yellow, orange, and red as the perceived risk increases, reflecting heightened passenger vigilance and cognitive engagement.

Using this method, we observe similar cognitive responses in other road events, such as pedestrian crossing (Events 1--4) and lead vehicle emergency braking (Events 11--14), demonstrating a transition from Calm to Nervous and from passive observation to active risk assessment. However, interesting variations exist across multiple tasks, reflecting differences in cognition and reaction times.

In Event 2 (Pedestrian crossing from the right sidewalk), \ac{EEG} topographic maps reveal a distinct P300 response at approximately 300\,ms, indicating the passenger's recognition of the pedestrian's movement. In contrast, for Event 1 (Pedestrian crossing from the left sidewalk), the P300 response is delayed to approximately 400\,ms. This latency difference arises because the vehicle is driving on the right side, making it easier for passengers to detect pedestrians approaching from the right, while those on the left are harder to observe and take longer to enter the vigilance zone.

We also observe that events involving leading vehicles in the ego lane (Events 11--14) show delayed or weaker P300 components compared to other scenarios. This is likely because these events are harder to detect---unlike sudden pedestrian crossings or cut-ins, they lack a clear triggering action. Individual differences in risk perception play a role, as some passengers are more tolerant of short following distances. When averaging \ac{EEG} data to generate topographic maps, this variation in perception results in a weaker overall response.

\subsubsection{Passenger Cognitive Model (PCM)}
\label{sec:pcm-model}
\begin{figure}[tbp]
\centering
\begin{minipage}[t]{0.485\textwidth}
    \vspace{0pt}
    \centering
    \captionsetup{font=footnotesize,skip=4pt}
    \parbox[c][0.22\textheight][c]{\linewidth}{\centering
        \includegraphics[width=\linewidth,height=0.22\textheight,keepaspectratio,trim=20 0 230 0,clip]{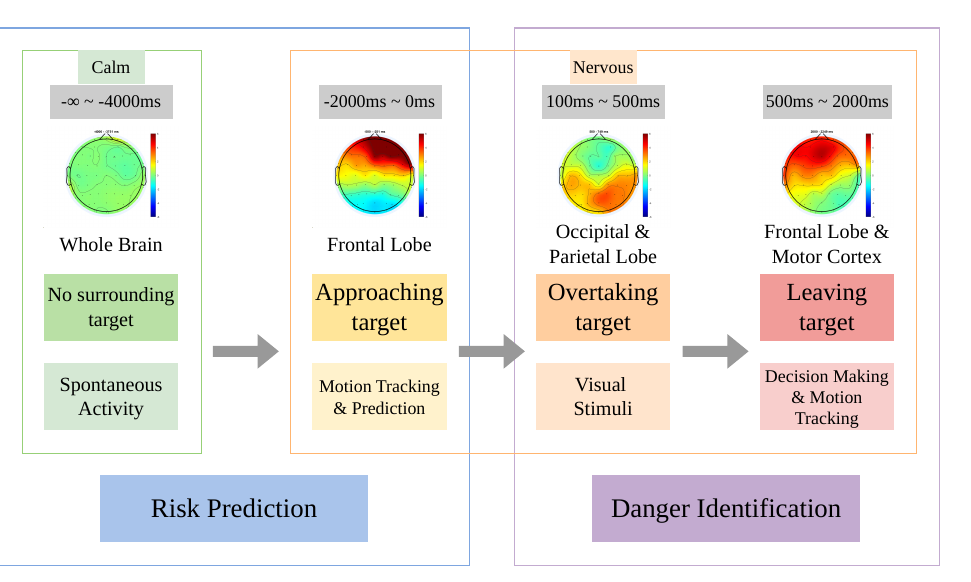}}
    \captionof{figure}{The Passenger Cognitive Model (\ac{PCM}) illustrates ``Calm'' and ``Nervous'' mental states according to surrounding traffic scenarios and topographic plots and summarizes the overall cognitive process.}
    \label{fig:pcm-model}
\end{minipage}\hfill
\begin{minipage}[t]{0.485\textwidth}
    \vspace{0pt}
    \centering
    \captionsetup{font=footnotesize,skip=4pt}
    \parbox[c][0.22\textheight][c]{\linewidth}{\centering
        \includegraphics[width=\linewidth,height=0.22\textheight,keepaspectratio]{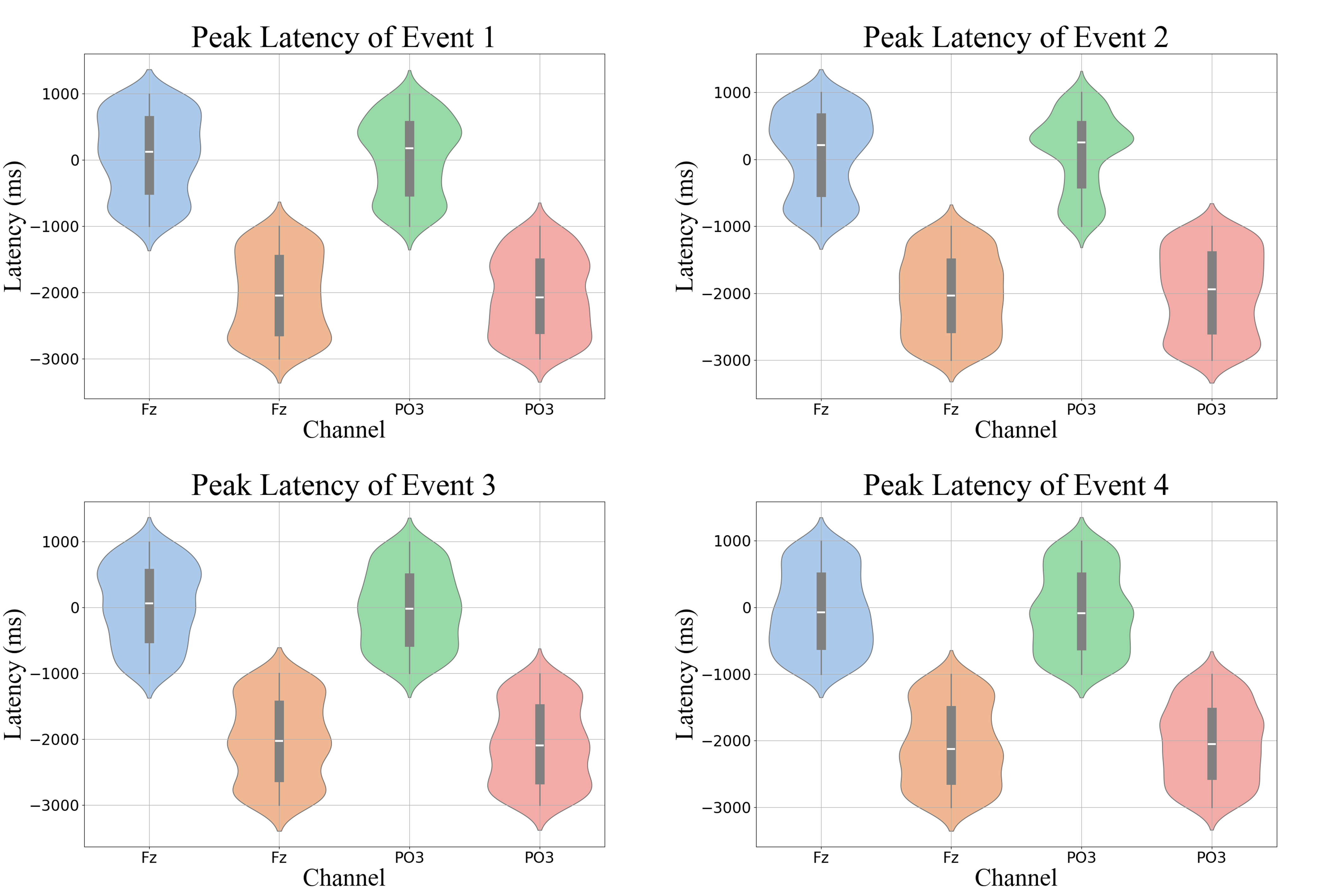}}
    \captionof{figure}{Peak Latency Analysis of Channels ``Fz'' and ``PO3'' under Event 1 (top left), Event 2 (bottom left), Event 3 (top right), and Event 4 (bottom right).}
    \label{fig:peak-latency-comparisons}
\end{minipage}
\end{figure}

Despite slight variations in response time, consistent patterns can be summarized in passenger cognition. Based on the established connection between traffic scenarios and active brain regions, we introduce the \ac{PCM} as illustrated in Fig.~\ref{fig:pcm-model}.

\ac{PCM} reveals the neural mechanisms of passengers while traveling in \acp{AV}, including aspects such as risk prediction, hazard perception, and decision-making capabilities. We will explain this model from seven perspectives, from top to bottom: mental states, time steps, topographic maps, active regions, traffic scenarios, cognition, and classification tasks that we formulate from \ac{PCM}.

\textbf{Calm State -- } Starting with traffic scenarios, ``Navigating with no surrounding targets'' is characterized by the absence of any other visible traffic participants or the presence of distant targets, posing no immediate potential hazards (time-to-event $>$ 5\,s). Under such circumstances, passengers exhibit a state of calmness, labeled ``Calm'' in Fig.~\ref{fig:pcm-model}, as evidenced by the topographic representation of brain activity, which displays an absence of specific brain region activity, resembling a state of spontaneous activity.

\textbf{Transition 1 --} As the ego vehicle approaches other traffic participants, entering the ``Approaching Target'' phase, the situation becomes riskier, leading passengers to experience heightened nervousness.

\textbf{Nervous State --} Passengers begin actively observing nearby vehicles and mentally simulating potential hazards. Starting 2\,s before event onset, this cognitive engagement triggers activity in the frontal lobes, associated with the predictive function~\cite{Miller_2000}, reinforcing the risk assessment process, and in the frontal eye field, which controls voluntary eye movement~\cite{Pouget_2015}.

\textbf{Transition 2 --} When the ego vehicle interacts with targets, e.g., during overtaking, it may encounter hazardous situations, such as abrupt vehicle cut-in or sudden crossing of pedestrians, depending on the event risk level, and the passengers are exposed to relevant visual stimuli. This leads to significant activity in the occipital and parietal lobes, the regions of the brain associated with visual processing and coordination~\cite{Culham_Valyear_2006}, as visually depicted in the topographic maps at 100--500\,ms. This notable feature corresponds to the P300 component, which occurs approximately 300\,ms after event onset.

\textbf{Decision-Making --} Passengers tend to intervene to prevent upcoming threats. This intention to intervene~\cite{Culham_Valyear_2006} is marked by heightened activity in the frontal lobes and potentially the motor cortex if the passenger imagines the action of physical intervention, as evident in the topographic maps. This step usually occurs immediately after the P300 components emerge.

\textbf{Task Settings --} \ac{PCM} captures two main transitions: from the Calm state to the Nervous state and from active observation to decision-making. These transitions provide critical insights into passengers' cognitive processes of identifying potential risks and discerning specific hazardous actions. Accordingly, we define the decoding tasks as \ac{RP} and \ac{DI}.

\subsection{Passenger EEG Decoding Strategy (PEDS)}
In this section, we formulate the \ac{PEDS} to answer the second question---\textit{How can we formulate the decoding task to fully leverage these cognitive insights for \ac{AV} safety?} \S~\ref{sec:peds-risk-level-definition} defines the Risky and Dangerous states and organizes the data into Safe, Low-Risk, and High-Risk groups. \S~\ref{sec:peds-rsl-labeling} applies \ac{RSL} to derive coupled labels for \ac{RP} and \ac{DI}. \S~\ref{sec:time-window-selection} uses peak-latency analysis to determine the task-specific \ac{EEG} windows. Finally, \S~\ref{sec:peds-decoding-model} presents the task-coupled \ac{3D-CRNN} that implements the proposed decoding strategy.

\subsubsection{\textbf{Categorization of EEG Data -- Risk Level Definition}}
\label{sec:peds-risk-level-definition}
In our problem formulation, we define two risk-level terms: Risky (Definition~\ref{def:risky}) and Dangerous (Definition~\ref{def:dangerous}). The Risky term quantifies the presence of surrounding traffic participants, while the Dangerous term assesses the occurrence of specific hazardous behaviors. Based on these two risk levels, we categorize the data into three groups: Safe, Low-Risk, and High-Risk. In brief, Safe corresponds to the absence of nearby risk-relevant targets, Low-Risk corresponds to the presence of surrounding targets without an immediate hazard, and High-Risk corresponds to situations in which an immediate hazard is present. These groups provide the semantic basis for the sequential labeling strategy and the two binary decoding tasks defined below.

\subsubsection{\textbf{Data Labeling -- RSL}}
\label{sec:peds-rsl-labeling}
For both event-related and event-free data, we apply \ac{RSL} by assigning two distinct labels to different time windows of a single data segment. This allows the model to perform \ac{RP} and \ac{DI} simultaneously.

\ac{RP} aims to assess potential risk during the pre-event stage by determining whether the current traffic situation is Risky or Not Risky. Under this scheme, both Low-Risk and High-Risk samples are labeled as Risky, while Safe samples are labeled as Not Risky. 

\ac{DI} focuses on evaluating hazardous behaviors. Here, High-Risk events are classified as Dangerous, whereas Low-Risk and Safe events are categorized as Not Dangerous (Definition~\ref{def:dangerous}).

Together, these labels define a risk-aware sequential framework that reflects cognitive responses across different stages of hazard evolution. Since \ac{RP} captures earlier contextual risk information, its encoded features may further assist \ac{DI}.

\subsubsection{\textbf{Time Window Selection -- Peak Latency Analysis}}
\label{sec:time-window-selection}
After setting up the classification tasks and data labeling, we need to determine appropriate time windows that capture sufficient discriminative features while meeting timing constraints. Specifically, candidates for \ac{RP} should rely solely on data preceding the event onset, with a preference for a longer gap between the end of the time window and event onset to ensure the prediction is made in advance without being influenced by the event itself. For \ac{DI}, the selected time window should enable a timely response, meaning it should be as close to the event onset as possible while still allowing the model to accurately detect hazardous behavior. This minimizes latency and ensures that the system can react promptly to emerging dangers.

The topographic plots provide direct visualization of brain activity, yet lack more intuitive and consistent evidence of which time point contains the most distinct and prominent features. Hence, we introduce peak latency analysis, a process of identifying and analyzing the timing of specific peaks in the \ac{EEG} signal. These peaks are usually associated with intense neural events, such as sensory, cognitive, or motor processes, providing valuable insights into the temporal dynamics of brain activity under selected traffic events. By inspecting the latency distribution, we can determine the most suitable window for the \ac{EEG} classification tasks.

Fig.~\ref{fig:peak-latency-comparisons} illustrates the peak latency of two key \ac{EEG} channels: Fz, situated in the frontal lobe and linked to executive functioning and attention regulation, and PO3, positioned in the parietal-occipital region, which plays a crucial role in visual processing and spatial awareness.

These comparisons, conducted across all subjects and Events 1--4 (Table~\ref{tab:events_table}), ensure generalization while accounting for inter-event and inter-subject variability. They not only guide optimal time-window selection but also highlight differences in event urgency. Events 1 and 2 correspond to High-Risk scenarios, while Events 3 and 4 represent Low-Risk scenarios. Notably, Event 2, with a shorter trigger distance (55\,m vs. 120\,m for Event 1), is more urgent and easier to detect.

For \ac{RP}, the orange and red violin plots (pre-event data) show consistent risk factors across events, with peak latencies primarily from -2500\,ms to -100\,ms, defining the time window as \([-2\,\mathrm{s},\,-0.5\,\mathrm{s}]\). For \ac{DI}, the blue and green violin plots (post-event data) reveal concentrated peak latencies around 300\,ms in the PO3 channel, particularly in Event 2, which has lower variability due to its immediacy. This leads to a \ac{DI} time window of \([-0.2\,\mathrm{s},\,1\,\mathrm{s}]\), covering peak latencies from 0 to 0.7\,s.

\subsubsection{\textbf{Decoding Model Construction -- 3D-CRNN}}
\label{sec:peds-decoding-model}
\begin{figure}[tbp]
    \centering
    \includegraphics[trim=0 450 0 0, clip, width=\textwidth]{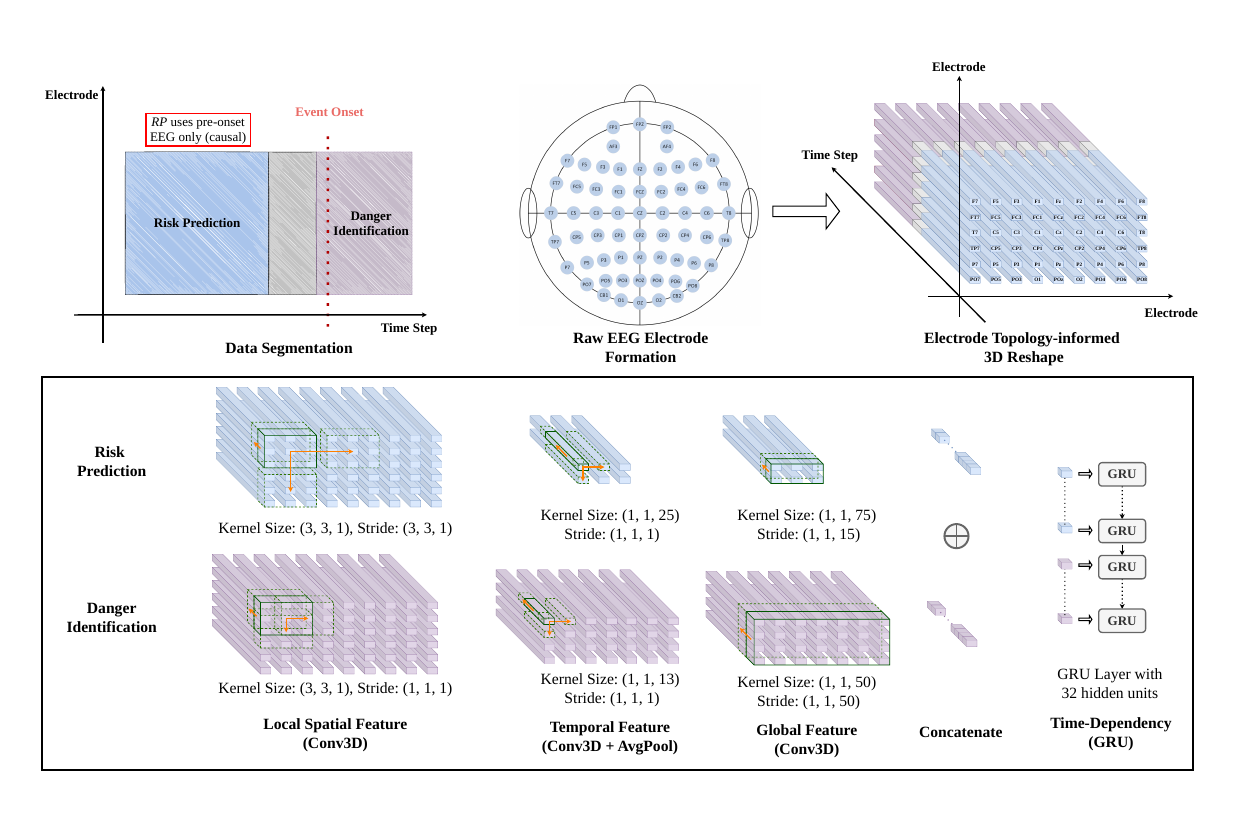}
    \caption{Architecture of the proposed \ac{3D-CRNN}. The input \ac{EEG} is first segmented into a causal pre-onset window for \ac{RP} and an event-centered window for \ac{DI}. Each branch reshapes the \ac{EEG} into an electrode topology-informed 3D representation and extracts features through local spatial convolution, temporal convolution/pooling, and global convolution. The branch features are then concatenated and modeled by a GRU layer with 32 hidden units for final classification. Blue denotes the \ac{RP} branch and purple denotes the \ac{DI} branch.}
    \label{fig:crnn}
\end{figure}

The core of \ac{PEDS} is a task-coupled \ac{3D-CRNN} for \ac{EEG} decoding. The network is optimized end-to-end with supervision signals from both \ac{RP} and \ac{DI}. As shown in Fig.~\ref{fig:crnn}, based on the task-specific window determined by peak latency analysis, the network first segments the input \ac{EEG} into two branches: a causal pre-onset branch for \ac{RP} and an event-centered branch for \ac{DI}, then maps each branch to an electrode topology-informed 3D representation, extracts branch-wise spatial--temporal features through 3D convolutions, and finally fuses the two feature streams with a GRU for prediction.

Let the input \ac{EEG} segment be denoted by
$
X \in \mathbb{R}^{C \times T},
$
where \(C\) is the number of electrodes and \(T\) is the number of temporal samples. In our setting, \(C=54\), the sampling rate is \(250\,\mathrm{Hz}\), and the full interval \([-2\,\mathrm{s},\,1\,\mathrm{s}]\) contains \(T=750\) samples. Following the selected task windows, the input is first segmented into two task-specific matrices,
\[
X_{\mathcal{RP}} \in \mathbb{R}^{C \times T_{\mathcal{RP}}}, \qquad
X_{\mathcal{DI}} \in \mathbb{R}^{C \times T_{\mathcal{DI}}},
\]
where \(T_{\mathcal{RP}} = 375\) and \(T_{\mathcal{DI}} = 300\). Specifically,
\[
X_{\mathcal{RP}} = X[:,\,1\!:\!375], \qquad
X_{\mathcal{DI}} = X[:,\,451\!:\!750].
\]

Thus, \(X_{\mathcal{RP}}\) covers the causal pre-onset interval \([-2\,\mathrm{s},\,-0.5\,\mathrm{s}]\), while \(X_{\mathcal{DI}}\) covers the event-centered interval \([-0.2\,\mathrm{s},\,1\,\mathrm{s}]\). Importantly, the \ac{RP} branch uses \textbf{only pre-onset \ac{EEG}}, thereby preserving causality in the prediction task.

To exploit the spatial topology of intrinsic neural activity, each segmented input is rearranged from the raw channel order into an electrode topology-informed 3D tensor. Let \(\Phi(\cdot)\) denote the electrode mapping from the original channel sequence to a \(6 \times 9\) spatial layout. Then, for \(b \in \{\mathcal{RP}, \mathcal{DI}\}\),
\[
\widetilde{X}_b = \Phi(X_b) \in \mathbb{R}^{H \times W \times T_b \times 1},
\: (H,W) = (6,9).
\]
This topology-informed reshape preserves neighborhood relationships among electrodes and allows the subsequent 3D convolutions to jointly model spatial structure and temporal dynamics.

The model contains two parallel CNN encoder branches. Although they share the same three-stage structure---local spatial feature extraction, temporal feature extraction, and global feature integration---the specific parameters differ to suit each task:
\begin{itemize}
    \item \textbf{3D-CRNN (\ac{RP})}: the encoder for Risk Prediction, operating on the causal pre-onset segment \(X_{\mathcal{RP}}\). Since this task relies on a longer observation window and gradually evolving anticipatory activity, the branch uses coarser early spatial downsampling with stride \((3,3,1)\), longer temporal kernels \((1,1,25)\) and \((1,1,75)\), and a moderate temporal stride \((1,1,15)\) to capture broad and slowly varying contextual information.
    
    \item \textbf{3D-CRNN (\ac{DI})}: the encoder for \ac{DI}, operating on the event-centered segment \(X_{\mathcal{DI}}\). Since this task targets more stimulus-locked and temporally precise responses, the branch preserves denser local spatial sampling with stride \((1,1,1)\), employs shorter temporal kernels \((1,1,13)\) and \((1,1,50)\), and uses stronger temporal compression with stride \((1,1,50)\) to summarize compact event-related responses.
\end{itemize}

Following \ac{EEG} decoding designs inspired by \ac{FBCSP}~\cite{Ang2012} and ShallowConvNet~\cite{DeepLearningEEG}, we adopt the following design choices for the temporal-feature stage:
\begin{enumerate}
    \item \textbf{Squaring nonlinearity} to emphasize high-energy \ac{EEG} components.
    \item \textbf{Average pooling} to reduce dimensionality while retaining important patterns.
    \item \textbf{Logarithmic activation} for enhancing separability between different risk levels.
\end{enumerate}

Let \(f_{\mathcal{RP}}(\cdot;\theta_{\mathcal{RP}})\) and \(f_{\mathcal{DI}}(\cdot;\theta_{\mathcal{DI}})\) denote the two branch encoders. Their outputs are
\begin{equation}
    Z_{\mathcal{RP}} = f_{\mathcal{RP}}(\widetilde{X}_{\mathcal{RP}};\theta_{\mathcal{RP}})
\end{equation}

\begin{equation}
Z_{\mathcal{DI}} = f_{\mathcal{DI}}(\widetilde{X}_{\mathcal{DI}};\theta_{\mathcal{DI}}).
\end{equation}
After permutation to align the temporal dimension, the branch-wise feature sequences are concatenated along the temporal axis:
\begin{equation}
Z = \mathrm{Concat}\!\left(Z_{\mathcal{RP}}, Z_{\mathcal{DI}}\right).
\end{equation}
The fused representation \(Z\) is then fed into a GRU layer with 32 hidden units,
\begin{equation}
H = \mathrm{GRU}(Z;\theta_g),
\end{equation}
to model temporal dependencies across the branch-wise encoded features. Overall, the network produces three outputs: a branch-wise \ac{RP} output, a branch-wise \ac{DI} output, and a fused output obtained after concatenation and GRU-based temporal modeling, which serves as the final \ac{DI} prediction with \ac{RSL} information incorporated through feature fusion.

Specifically, each CNN encoder is followed by a lightweight Dense classification head with Softmax activation, and the fused representation is passed through a dropout layer with a dropout rate of \(0.5\) and a final classifier. The class posterior is given by
\begin{equation}
\hat{y}_{k} = \frac{\exp(z_k)}{\sum_{j=1}^{K}\exp(z_j)}, \qquad k=1,\dots,K,
\end{equation}
where \(K\) is the number of classes. In this study, both \ac{RP} and \ac{DI} are binary tasks, and thus \(K=2\).

To mitigate class imbalance, we use a weighted cross-entropy loss. For a task with \(N\) samples and \(K\) classes, the loss is defined as
\begin{equation}
\mathcal{L}
=
-\frac{1}{N}
\sum_{n=1}^{N}
\sum_{k=1}^{K}
w_k\, y_{n,k}\log \hat{y}_{n,k},
\label{eq:wce}
\end{equation}
where \(y_{n,k}\in\{0,1\}\) is the one-hot ground-truth label, \(\hat{y}_{n,k}\) is the predicted posterior probability, and \(w_k\) is the class weight. 

During training, \ac{RP} supervision is used to optimize the \ac{RP} branch, while \ac{DI} supervision is used to optimize the \ac{DI} branch separately. The fused representation is then further optimized end-to-end using \ac{DI} labels. In this way, the final \ac{DI} prediction benefits from both branch-specific feature learning and temporal feature fusion.

\section{Results}
\label{sec:Results}

The \ac{EEG} decoding results are presented step by step by addressing the following key questions:
\begin{itemize}
    \item \textbf{Can we predict risk before event onset using \ac{PEDS}?}
    \\ Evaluating whether the pre-event \ac{EEG} window contains sufficient discriminative information for \ac{RP} across all compared models.
    \item \textbf{Does \ac{RSL} improve performance in \ac{DI}?}
    \\ Assessing whether \ac{RSL}, which incorporates \ac{RP} information into \ac{DI}, improves hazardous-event decoding.
    \item \textbf{Are the gains consistent across event types?}
    \\ Comparing model behavior on \ac{AEB}, cut-in, and pedestrian scenarios to examine whether the proposed framework remains effective under different hazard mechanisms.
    \item \textbf{How well does \ac{PEDS} generalize across sessions?}
    \\ Investigating intra-subject variability by testing whether models trained on one group of sessions remain effective on a held-out session recorded on a different day.
    \item \textbf{How well does \ac{PEDS} generalize across subjects?}
    \\ Investigating inter-subject variability by evaluating seen-subject and unseen-subject performance under a leave-one-subject-out protocol.
\end{itemize}
\S~\ref{res:training-details} and \S~\ref{res:metrics} list the training details and evaluation metrics.
\S~\ref{res:risk-prediction} evaluates \ac{RP} across all models.
\S~\ref{res:single-subject-di} examines the \ac{RSL} ablation and single-subject \ac{DI} performance.
\S~\ref{res:Eventwise} reports event-wise comparisons on \ac{AEB}, cut-in, and pedestrian scenarios.
\S~\ref{res:cross-session} evaluates cross-session generalization for intra-subject variability.
\S~\ref{res:cross-subject-di} evaluates cross-subject generalization for inter-subject variability.
Finally, \S~\ref{res:online} reports model complexity and inference latency for potential online deployment.

\subsection{Training Details}
\label{res:training-details}

The \ac{3D-CRNN}, EEGNet~\cite{EEGNet}, ShallowConvNet~\cite{DeepLearningEEG}, DeepConvNet~\cite{DeepLearningEEG}, CTNet~\cite{Zhao2024CTNet}, and DSC-ConvLSTM~\cite{DSCConvLSTM} were implemented in TensorFlow, whereas XGB-DIM~\cite{bib8} was implemented in PyTorch. Data were split into 70\% for training and 30\% for testing. During training, 30\% of the training set was further used for validation. More details, including sample numbers per split, session usage, and fold assignments, are presented in Appendix~\ref{app:data-split}.

The number of epochs was 100, the batch size was 32, and the optimizer was Adam with a learning rate of \(10^{-4}\). Early stopping monitored the validation loss with a patience of 10 epochs, and the best model was saved according to the minimum validation loss.

\subsection{Evaluation Metrics}
\label{res:metrics}

We report Precision, Recall, F1-score, and \ac{BA} to evaluate classification performance. Since the class distribution is not perfectly balanced, \ac{BA}, the macro-averaged metric across classes, is used as the primary metric. For binary classification, with \(TP\), \(TN\), \(FP\), and \(FN\) denoting true positives, true negatives, false positives, and false negatives, respectively,
\noindent\begin{minipage}[t]{0.485\textwidth}
\begin{equation}
\mathrm{Precision} = \frac{TP}{TP+FP},
\label{eq:precision}
\end{equation}
\end{minipage}\hfill
\begin{minipage}[t]{0.485\textwidth}
\begin{equation}
\mathrm{Recall} = \frac{TP}{TP+FN},
\label{eq:recall}
\end{equation}
\end{minipage}

\noindent\begin{minipage}[t]{0.485\textwidth}
\begin{equation}
\mathrm{F1\text{-}score} = \frac{2TP}{2TP+FP+FN},
\label{eq:f1}
\end{equation}
\end{minipage}\hfill
\begin{minipage}[t]{0.485\textwidth}
\begin{equation}
\mathrm{BA} = \frac{1}{2}\left(\frac{TP}{TP+FN} + \frac{TN}{TN+FP}\right).
\label{eq:ba}
\end{equation}
\end{minipage}

Unless otherwise specified, the reported results are averaged across subjects and presented as mean \(\pm\) standard deviation. \textbf{Best} results are in bold and \underline{second-best} results are underlined within each dataset.

\begin{table}[tbp]
\footnotesize
\centering
\setlength{\tabcolsep}{\dimexpr(\textwidth-13.3cm)/12\relax}
\begin{tabular}{
>{\centering\arraybackslash}p{2cm}
>{\centering\arraybackslash}p{2.5cm}
>{\centering\arraybackslash}p{2cm}
>{\centering\arraybackslash}p{2cm}
>{\centering\arraybackslash}p{2cm}
>{\centering\arraybackslash}p{2.8cm}
}
\toprule[1.5pt]
Setting & Model & Precision $\uparrow$ & Recall $\uparrow$ & F1-score $\uparrow$ & Balanced Accuracy $\uparrow$ \\
\midrule
\multirow{7}{*}{RP}
& EEGNet & 0.872 $\pm$ 0.129 & 0.864 $\pm$ 0.129 & 0.863 $\pm$ 0.130 & 0.860 $\pm$ 0.129 \\
& ShallowConvNet & \underline{0.953 $\pm$ 0.028} & \underline{0.950 $\pm$ 0.031} & \underline{0.949 $\pm$ 0.031} & \underline{0.947 $\pm$ 0.032} \\
& DeepConvNet & 0.727 $\pm$ 0.125 & 0.649 $\pm$ 0.146 & 0.602 $\pm$ 0.187 & 0.646 $\pm$ 0.142 \\
& XGB-DIM & 0.554 $\pm$ 0.053 & 0.552 $\pm$ 0.054 & 0.552 $\pm$ 0.054 & 0.553 $\pm$ 0.053 \\
& CTNet & 0.734 $\pm$ 0.125 & 0.726 $\pm$ 0.126 & 0.719 $\pm$ 0.130 & 0.719 $\pm$ 0.130 \\
& DSC-ConvLSTM & 0.722 $\pm$ 0.125 & 0.708 $\pm$ 0.138 & 0.697 $\pm$ 0.154 & 0.710 $\pm$ 0.129 \\
\rowcolor{AUINHighlight}
& 3D-CRNN & \textbf{0.957 $\pm$ 0.026} & \textbf{0.955 $\pm$ 0.026} & \textbf{0.955 $\pm$ 0.027} & \textbf{0.953 $\pm$ 0.027} \\
\bottomrule
\end{tabular}

\vspace{2pt}
\caption{Subject-wise performance of models on Risk Prediction.}
\label{tab:performance_rp}
\end{table}

\begin{table}[tbp]
\footnotesize
\centering
\setlength{\tabcolsep}{\dimexpr(\textwidth-13.3cm)/12\relax}
\begin{tabular}{
>{\centering\arraybackslash}p{2cm}
>{\centering\arraybackslash}p{2.5cm}
>{\centering\arraybackslash}p{2cm}
>{\centering\arraybackslash}p{2cm}
>{\centering\arraybackslash}p{2cm}
>{\centering\arraybackslash}p{2.8cm}
}
\toprule[1.5pt]
Setting & Model & Precision $\uparrow$ & Recall $\uparrow$ & F1-score $\uparrow$ & Balanced Accuracy $\uparrow$ \\
\midrule
\multirow{7}{*}{DI without RSL} & EEGNet & \underline{0.813 $\pm$ 0.062} & \underline{0.783 $\pm$ 0.060} & \underline{0.787 $\pm$ 0.060} & \underline{0.782 $\pm$ 0.076} \\
 & ShallowConvNet & 0.805 $\pm$ 0.034 & 0.771 $\pm$ 0.037 & 0.777 $\pm$ 0.036 & 0.778 $\pm$ 0.046 \\
 & DeepConvNet & 0.745 $\pm$ 0.074 & 0.655 $\pm$ 0.104 & 0.666 $\pm$ 0.102 & 0.688 $\pm$ 0.097 \\
 & XGB-DIM & 0.685 $\pm$ 0.022 & 0.569 $\pm$ 0.059 & 0.600 $\pm$ 0.051 & 0.564 $\pm$ 0.041 \\
 & CTNet & 0.682 $\pm$ 0.090 & 0.610 $\pm$ 0.094 & 0.615 $\pm$ 0.093 & 0.611 $\pm$ 0.106 \\
 & DSC-ConvLSTM & 0.775 $\pm$ 0.051 & 0.712 $\pm$ 0.076 & 0.723 $\pm$ 0.073 & 0.736 $\pm$ 0.067 \\
\rowcolor{AUINHighlight}
 & 3D-CRNN & \textbf{0.833 $\pm$ 0.030} & \textbf{0.811 $\pm$ 0.030} & \textbf{0.814 $\pm$ 0.032} & \textbf{0.809 $\pm$ 0.039} \\
\midrule
\multirow{7}{*}{DI with RSL} & EEGNet & 0.829 $\pm$ 0.028 & 0.820 $\pm$ 0.024 & 0.821 $\pm$ 0.024 & \underline{0.803 $\pm$ 0.043} \\
 & ShallowConvNet & 0.828 $\pm$ 0.020 & 0.816 $\pm$ 0.015 & 0.817 $\pm$ 0.016 & \underline{0.803 $\pm$ 0.028} \\
 & DeepConvNet & 0.815 $\pm$ 0.042 & 0.806 $\pm$ 0.030 & 0.804 $\pm$ 0.035 & 0.776 $\pm$ 0.066 \\
 & XGB-DIM & \underline{0.831 $\pm$ 0.017} & \textbf{0.838 $\pm$ 0.015} & \underline{0.830 $\pm$ 0.017} & 0.734 $\pm$ 0.027 \\
 & CTNet & 0.737 $\pm$ 0.098 & 0.759 $\pm$ 0.042 & 0.735 $\pm$ 0.078 & 0.682 $\pm$ 0.107 \\
 & DSC-ConvLSTM & 0.772 $\pm$ 0.044 & 0.773 $\pm$ 0.034 & 0.768 $\pm$ 0.042 & 0.722 $\pm$ 0.068 \\
\rowcolor{AUINHighlight}
 & 3D-CRNN & \textbf{0.862 $\pm$ 0.026} & \underline{0.829 $\pm$ 0.023} & \textbf{0.835 $\pm$ 0.023} & \textbf{0.850 $\pm$ 0.032} \\
\bottomrule
\end{tabular}

\vspace{2pt}
\caption{Subject-wise performance comparison of models with and without \ac{RSL} on \ac{DI}.}
\label{tab:performance_rsl}
\end{table}

\begin{table}[tbp]
\footnotesize
\centering
\setlength{\tabcolsep}{\dimexpr(\textwidth-13.3cm)/12\relax}
\begin{tabular}{
>{\centering\arraybackslash}p{2cm}
>{\centering\arraybackslash}p{2.5cm}
>{\centering\arraybackslash}p{2cm}
>{\centering\arraybackslash}p{2cm}
>{\centering\arraybackslash}p{2cm}
>{\centering\arraybackslash}p{2.8cm}
}
\toprule[1.5pt]
Setting & Model & Precision $\uparrow$ & Recall $\uparrow$ & F1-score $\uparrow$ & Balanced Accuracy $\uparrow$ \\
\midrule
\multirow{7}{*}{RP}
& EEGNet & 0.871 $\pm$ 0.076 & 0.884 $\pm$ 0.043 & 0.857 $\pm$ 0.064 & 0.676 $\pm$ 0.133 \\
& ShallowConvNet & \underline{0.939 $\pm$ 0.073} & \underline{0.943 $\pm$ 0.059} & \underline{0.938 $\pm$ 0.068} & \underline{0.857 $\pm$ 0.137} \\
& DeepConvNet & 0.766 $\pm$ 0.091 & 0.833 $\pm$ 0.031 & 0.772 $\pm$ 0.035 & 0.517 $\pm$ 0.036 \\
& XGB-DIM & 0.748 $\pm$ 0.037 & 0.528 $\pm$ 0.060 & 0.593 $\pm$ 0.050 & 0.509 $\pm$ 0.066 \\
& CTNet & 0.777 $\pm$ 0.066 & 0.829 $\pm$ 0.029 & 0.781 $\pm$ 0.029 & 0.537 $\pm$ 0.042 \\
& DSC-ConvLSTM & 0.810 $\pm$ 0.072 & 0.792 $\pm$ 0.172 & 0.756 $\pm$ 0.168 & 0.576 $\pm$ 0.071 \\
\rowcolor{AUINHighlight}
& 3D-CRNN & \textbf{0.961 $\pm$ 0.035} & \textbf{0.958 $\pm$ 0.042} & \textbf{0.951 $\pm$ 0.061} & \textbf{0.880 $\pm$ 0.127} \\
\midrule
\multirow{7}{*}{DI without RSL}
& EEGNet & 0.878 $\pm$ 0.026 & 0.908 $\pm$ 0.011 & 0.883 $\pm$ 0.012 & 0.552 $\pm$ 0.048 \\
& ShallowConvNet & \underline{0.890 $\pm$ 0.031} & 0.908 $\pm$ 0.018 & \underline{0.897 $\pm$ 0.025} & 0.627 $\pm$ 0.091 \\
& DeepConvNet & 0.836 $\pm$ 0.018 & \textbf{0.912 $\pm$ 0.006} & 0.872 $\pm$ 0.011 & 0.503 $\pm$ 0.011 \\
& XGB-DIM & 0.888 $\pm$ 0.026 & 0.667 $\pm$ 0.077 & 0.740 $\pm$ 0.061 & \underline{0.650 $\pm$ 0.097} \\
& CTNet & 0.857 $\pm$ 0.027 & 0.880 $\pm$ 0.033 & 0.867 $\pm$ 0.024 & 0.549 $\pm$ 0.085 \\
& DSC-ConvLSTM & 0.866 $\pm$ 0.034 & 0.862 $\pm$ 0.080 & 0.860 $\pm$ 0.054 & 0.579 $\pm$ 0.092 \\
\rowcolor{AUINHighlight}
& 3D-CRNN & \textbf{0.923 $\pm$ 0.039} & \underline{0.912 $\pm$ 0.049} & \textbf{0.915 $\pm$ 0.043} & \textbf{0.776 $\pm$ 0.130} \\
\midrule
\multirow{7}{*}{DI with RSL}
& EEGNet & 0.851 $\pm$ 0.034 & 0.913 $\pm$ 0.009 & 0.876 $\pm$ 0.013 & 0.513 $\pm$ 0.025 \\
& ShallowConvNet & \underline{0.896 $\pm$ 0.032} & \underline{0.917 $\pm$ 0.015} & \underline{0.902 $\pm$ 0.023} & \underline{0.626 $\pm$ 0.086} \\
& DeepConvNet & 0.833 $\pm$ 0.012 & 0.913 $\pm$ 0.007 & 0.871 $\pm$ 0.010 & 0.500 $\pm$ 0.000 \\
& XGB-DIM & 0.877 $\pm$ 0.024 & 0.807 $\pm$ 0.059 & 0.836 $\pm$ 0.041 & 0.602 $\pm$ 0.094 \\
& CTNet & 0.838 $\pm$ 0.019 & 0.912 $\pm$ 0.007 & 0.873 $\pm$ 0.011 & 0.509 $\pm$ 0.030 \\
& DSC-ConvLSTM & 0.839 $\pm$ 0.018 & 0.910 $\pm$ 0.009 & 0.873 $\pm$ 0.010 & 0.510 $\pm$ 0.038 \\
\rowcolor{AUINHighlight}
& 3D-CRNN & \textbf{0.922 $\pm$ 0.037} & \textbf{0.922 $\pm$ 0.018} & \textbf{0.920 $\pm$ 0.026} & \textbf{0.778 $\pm$ 0.129} \\
\bottomrule
\end{tabular}
\caption{Event-wise performance comparison on \ac{AEB} scenarios.}
\label{tab:event_aeb}
\end{table}

\begin{table}[tbp]
\footnotesize
\centering
\setlength{\tabcolsep}{\dimexpr(\textwidth-13.3cm)/12\relax}
\begin{tabular}{
>{\centering\arraybackslash}p{2cm}
>{\centering\arraybackslash}p{2.5cm}
>{\centering\arraybackslash}p{2cm}
>{\centering\arraybackslash}p{2cm}
>{\centering\arraybackslash}p{2cm}
>{\centering\arraybackslash}p{2.8cm}
}
\toprule[1.5pt]
Setting & Model & Precision $\uparrow$ & Recall $\uparrow$ & F1-score $\uparrow$ & Balanced Accuracy $\uparrow$ \\
\midrule
\multirow{7}{*}{RP}
& EEGNet & 0.869 $\pm$ 0.095 & 0.870 $\pm$ 0.085 & 0.866 $\pm$ 0.095 & 0.840 $\pm$ 0.110 \\
& ShallowConvNet & \underline{0.955 $\pm$ 0.033} & \underline{0.953 $\pm$ 0.036} & \underline{0.952 $\pm$ 0.039} & \underline{0.933 $\pm$ 0.052} \\
& DeepConvNet & 0.678 $\pm$ 0.177 & 0.727 $\pm$ 0.129 & 0.681 $\pm$ 0.153 & 0.625 $\pm$ 0.171 \\
& XGB-DIM & 0.606 $\pm$ 0.048 & 0.537 $\pm$ 0.066 & 0.554 $\pm$ 0.063 & 0.537 $\pm$ 0.062 \\
& CTNet & 0.722 $\pm$ 0.112 & 0.739 $\pm$ 0.094 & 0.710 $\pm$ 0.116 & 0.652 $\pm$ 0.125 \\
& DSC-ConvLSTM & 0.789 $\pm$ 0.133 & 0.804 $\pm$ 0.082 & 0.777 $\pm$ 0.119 & 0.728 $\pm$ 0.123 \\
\rowcolor{AUINHighlight}
& 3D-CRNN & \textbf{0.963 $\pm$ 0.023} & \textbf{0.962 $\pm$ 0.024} & \textbf{0.961 $\pm$ 0.025} & \textbf{0.946 $\pm$ 0.032} \\
\midrule
\multirow{7}{*}{DI without RSL}
& EEGNet & 0.858 $\pm$ 0.033 & 0.859 $\pm$ 0.028 & 0.855 $\pm$ 0.039 & 0.789 $\pm$ 0.075 \\
& ShallowConvNet & \underline{0.864 $\pm$ 0.028} & \underline{0.865 $\pm$ 0.025} & \underline{0.863 $\pm$ 0.027} & \underline{0.797 $\pm$ 0.055} \\
& DeepConvNet & 0.711 $\pm$ 0.088 & 0.778 $\pm$ 0.022 & 0.704 $\pm$ 0.045 & 0.537 $\pm$ 0.055 \\
& XGB-DIM & 0.700 $\pm$ 0.041 & 0.602 $\pm$ 0.071 & 0.632 $\pm$ 0.063 & 0.574 $\pm$ 0.062 \\
& CTNet & 0.763 $\pm$ 0.087 & 0.720 $\pm$ 0.149 & 0.717 $\pm$ 0.159 & 0.670 $\pm$ 0.123 \\
& DSC-ConvLSTM & 0.818 $\pm$ 0.051 & 0.720 $\pm$ 0.161 & 0.727 $\pm$ 0.176 & 0.740 $\pm$ 0.098 \\
\rowcolor{AUINHighlight}
& 3D-CRNN & \textbf{0.896 $\pm$ 0.029} & \textbf{0.873 $\pm$ 0.037} & \textbf{0.879 $\pm$ 0.035} & \textbf{0.877 $\pm$ 0.042} \\
\midrule
\multirow{7}{*}{DI with RSL}
& EEGNet & 0.858 $\pm$ 0.036 & 0.862 $\pm$ 0.034 & 0.854 $\pm$ 0.048 & 0.772 $\pm$ 0.085 \\
& ShallowConvNet & \underline{0.871 $\pm$ 0.028} & \underline{0.871 $\pm$ 0.024} & \underline{0.870 $\pm$ 0.027} & \underline{0.813 $\pm$ 0.056} \\
& DeepConvNet & 0.688 $\pm$ 0.123 & 0.789 $\pm$ 0.039 & 0.713 $\pm$ 0.076 & 0.560 $\pm$ 0.121 \\
& XGB-DIM & 0.704 $\pm$ 0.033 & 0.657 $\pm$ 0.044 & 0.675 $\pm$ 0.038 & 0.580 $\pm$ 0.061 \\
& CTNet & 0.786 $\pm$ 0.083 & 0.811 $\pm$ 0.042 & 0.773 $\pm$ 0.072 & 0.643 $\pm$ 0.112 \\
& DSC-ConvLSTM & 0.812 $\pm$ 0.086 & 0.836 $\pm$ 0.039 & 0.819 $\pm$ 0.066 & 0.717 $\pm$ 0.101 \\
\rowcolor{AUINHighlight}
& 3D-CRNN & \textbf{0.901 $\pm$ 0.016} & \textbf{0.888 $\pm$ 0.013} & \textbf{0.892 $\pm$ 0.014} & \textbf{0.880 $\pm$ 0.026} \\
\bottomrule
\end{tabular}
\caption{Event-wise performance comparison on cut-in scenarios.}
\label{tab:event_cutin}
\end{table}

\begin{table}[tbp]
\footnotesize
\centering
\setlength{\tabcolsep}{\dimexpr(\textwidth-13.3cm)/12\relax}
\begin{tabular}{
>{\centering\arraybackslash}p{2cm}
>{\centering\arraybackslash}p{2.5cm}
>{\centering\arraybackslash}p{2cm}
>{\centering\arraybackslash}p{2cm}
>{\centering\arraybackslash}p{2cm}
>{\centering\arraybackslash}p{2.8cm}
}
\toprule[1.5pt]
Setting & Model & Precision $\uparrow$ & Recall $\uparrow$ & F1-score $\uparrow$ & Balanced Accuracy $\uparrow$ \\
\midrule
\multirow{7}{*}{RP}
& EEGNet & 0.865 $\pm$ 0.044 & 0.886 $\pm$ 0.036 & 0.862 $\pm$ 0.042 & 0.684 $\pm$ 0.088 \\
& ShallowConvNet & \underline{0.943 $\pm$ 0.053} & \underline{0.947 $\pm$ 0.046} & \underline{0.943 $\pm$ 0.051} & \underline{0.870 $\pm$ 0.095} \\
& DeepConvNet & 0.753 $\pm$ 0.051 & 0.833 $\pm$ 0.030 & 0.776 $\pm$ 0.034 & 0.539 $\pm$ 0.056 \\
& XGB-DIM & 0.752 $\pm$ 0.039 & 0.553 $\pm$ 0.032 & 0.612 $\pm$ 0.030 & 0.534 $\pm$ 0.048 \\
& CTNet & 0.766 $\pm$ 0.048 & 0.834 $\pm$ 0.017 & 0.790 $\pm$ 0.030 & 0.573 $\pm$ 0.064 \\
& DSC-ConvLSTM & 0.782 $\pm$ 0.113 & 0.800 $\pm$ 0.103 & 0.772 $\pm$ 0.118 & 0.608 $\pm$ 0.056 \\
\rowcolor{AUINHighlight}
& 3D-CRNN & \textbf{0.954 $\pm$ 0.059} & \textbf{0.961 $\pm$ 0.036} & \textbf{0.955 $\pm$ 0.050} & \textbf{0.889 $\pm$ 0.092} \\
\midrule
\multirow{7}{*}{DI without RSL}
& EEGNet & 0.872 $\pm$ 0.021 & \underline{0.906 $\pm$ 0.013} & 0.883 $\pm$ 0.014 & 0.582 $\pm$ 0.027 \\
& ShallowConvNet & \underline{0.894 $\pm$ 0.026} & \textbf{0.907 $\pm$ 0.018} & \underline{0.898 $\pm$ 0.023} & \underline{0.655 $\pm$ 0.070} \\
& DeepConvNet & 0.823 $\pm$ 0.021 & 0.904 $\pm$ 0.010 & 0.861 $\pm$ 0.015 & 0.508 $\pm$ 0.029 \\
& XGB-DIM & 0.871 $\pm$ 0.025 & 0.651 $\pm$ 0.047 & 0.723 $\pm$ 0.038 & 0.617 $\pm$ 0.063 \\
& CTNet & 0.862 $\pm$ 0.022 & 0.880 $\pm$ 0.028 & 0.864 $\pm$ 0.020 & 0.571 $\pm$ 0.069 \\
& DSC-ConvLSTM & 0.875 $\pm$ 0.038 & 0.863 $\pm$ 0.067 & 0.863 $\pm$ 0.052 & 0.625 $\pm$ 0.087 \\
\rowcolor{AUINHighlight}
& 3D-CRNN & \textbf{0.916 $\pm$ 0.027} & 0.903 $\pm$ 0.039 & \textbf{0.908 $\pm$ 0.033} & \textbf{0.760 $\pm$ 0.078} \\
\midrule
\multirow{7}{*}{DI with RSL}
& EEGNet & 0.861 $\pm$ 0.019 & 0.906 $\pm$ 0.008 & 0.873 $\pm$ 0.012 & 0.541 $\pm$ 0.030 \\
& ShallowConvNet & \underline{0.904 $\pm$ 0.032} & \underline{0.918 $\pm$ 0.019} & \underline{0.905 $\pm$ 0.025} & \underline{0.660 $\pm$ 0.073} \\
& DeepConvNet & 0.820 $\pm$ 0.016 & 0.903 $\pm$ 0.008 & 0.859 $\pm$ 0.011 & 0.504 $\pm$ 0.015 \\
& XGB-DIM & 0.869 $\pm$ 0.025 & 0.805 $\pm$ 0.033 & 0.830 $\pm$ 0.025 & 0.618 $\pm$ 0.064 \\
& CTNet & 0.830 $\pm$ 0.020 & 0.903 $\pm$ 0.008 & 0.862 $\pm$ 0.012 & 0.511 $\pm$ 0.018 \\
& DSC-ConvLSTM & 0.847 $\pm$ 0.031 & 0.907 $\pm$ 0.012 & 0.873 $\pm$ 0.021 & 0.544 $\pm$ 0.046 \\
\rowcolor{AUINHighlight}
& 3D-CRNN & \textbf{0.929 $\pm$ 0.020} & \textbf{0.925 $\pm$ 0.013} & \textbf{0.924 $\pm$ 0.018} & \textbf{0.799 $\pm$ 0.074} \\
\bottomrule
\end{tabular}
\caption{Event-wise performance comparison on pedestrian scenarios.}
\label{tab:event_ped}
\end{table}

\subsection{Risk Prediction}
\label{res:risk-prediction}

Table~\ref{tab:performance_rp} reports the subject-wise performance of all models on \ac{RP}. The \ac{RSL} module is designed for optimizing \ac{DI} using \ac{RP}, so the comparisons presented in this section are conducted without the \ac{RSL} module. The proposed \ac{3D-CRNN} achieves the best overall performance, with a Precision of \(0.957 \pm 0.026\), Recall of \(0.955 \pm 0.026\), F1-score of \(0.955 \pm 0.027\), and \ac{BA} of \(0.953 \pm 0.027\). These results show that the pre-event \ac{EEG} window contains discriminative neural patterns related to the presence of upcoming risk, allowing the model to produce an early warning before the explicit hazardous behavior occurs.

The strong performance of \ac{3D-CRNN} on \ac{RP} can be explained by the nature of the task. Unlike \ac{DI}, which is usually associated with a sudden stimulus after event onset, \ac{RP} does not contain a sharply time-locked external stimulus. Instead, it reflects a relatively long-term cognitive transition as passengers gradually perceive surrounding traffic participants and become more alert before a hazardous action occurs. Therefore, successful \ac{RP} requires the model to capture distributed spatio-temporal \ac{EEG} patterns over the pre-event interval rather than relying on a single precisely aligned evoked response.

Among the baselines, ShallowConvNet is the strongest competitor, reaching a \ac{BA} of \(0.947 \pm 0.032\). EEGNet also performs reliably with a \ac{BA} of \(0.860 \pm 0.129\), whereas DeepConvNet, CTNet, DSC-ConvLSTM, and XGB-DIM show lower \ac{BA}. In particular, XGB-DIM performs poorly for \ac{RP}, with a \ac{BA} of \(0.553 \pm 0.053\). This is consistent with its stronger dependence on precise event alignment and feature alignment: when the discriminative information is not a strong stimulus-locked response but a gradual cognitive-state change, handcrafted or strictly aligned features become less stable.

\subsection{Single-Subject Danger Identification}
\label{res:single-subject-di}

This experiment examines two aspects: the ability of each model to detect immediate hazardous behavior, i.e., \ac{DI}, and the contribution of \ac{RSL}.

For the \textbf{setting without \ac{RSL}}, models are trained directly on \ac{DI} labels only. For the \textbf{setting with \ac{RSL}}, \ac{RP} information is incorporated into \ac{DI} prediction. For baseline models that process a single target label, the \ac{RP} and \ac{DI} outputs are fused after separate training. For \ac{3D-CRNN}, \ac{RSL} is naturally embedded in the two-branch architecture, enabling the model to jointly learn pre-event risk and post-event danger representations.

As shown in Table~\ref{tab:performance_rsl}, \ac{3D-CRNN} is already the best model without \ac{RSL}, achieving a Precision of \(0.833 \pm 0.030\), Recall of \(0.811 \pm 0.030\), F1-score of \(0.814 \pm 0.032\), and \ac{BA} of \(0.809 \pm 0.039\). With \ac{RSL}, its performance further increases to a Precision of \(0.862 \pm 0.026\), Recall of \(0.829 \pm 0.023\), F1-score of \(0.835 \pm 0.023\), and \ac{BA} of \(0.850 \pm 0.032\). This corresponds to a \(0.041\) absolute improvement in \ac{BA}, indicating that pre-event risk information provides useful context for judging whether the subsequent event becomes dangerous.

\ac{RSL} also improves most baselines. EEGNet, ShallowConvNet, DeepConvNet, XGB-DIM, and CTNet all obtain higher \ac{BA} with \ac{RSL} than without \ac{RSL}. The most pronounced improvement occurs for XGB-DIM, whose Recall increases from \(0.569\) to \(0.838\) and whose F1-score increases from \(0.600\) to \(0.830\). However, its \ac{BA} remains lower than that of the CNN-based models, suggesting that the additional risk cue improves sensitivity but does not fully resolve class-balanced discrimination. Overall, the averaged single-subject results support both components of the proposed framework: \ac{3D-CRNN} provides the strongest \ac{EEG} representation, and \ac{RSL} improves \ac{DI} by using the sequential relationship between risk emergence and danger occurrence.

\subsection{Event-wise Analysis}
\label{res:Eventwise}

To further examine whether the proposed framework remains effective under different hazard mechanisms, we evaluate the models separately on \ac{AEB}, cut-in, and pedestrian scenarios, as shown in Tables~\ref{tab:event_aeb},~\ref{tab:event_cutin}, and~\ref{tab:event_ped}. These event-wise comparisons are important because different traffic events may elicit different passenger cognitive responses. For example, cut-in events involve lateral vehicle motion, pedestrian events involve vulnerable road users entering the driving path, and \ac{AEB} events involve the behavior of a leading vehicle.

For \ac{RP}, \ac{3D-CRNN} achieves the best performance across all three event types. It obtains \ac{BA} values of \(0.880 \pm 0.127\), \(0.946 \pm 0.032\), and \(0.889 \pm 0.092\) on \ac{AEB}, cut-in, and pedestrian scenarios, respectively. ShallowConvNet is consistently the second-best model, reaching \ac{BA} values of \(0.857 \pm 0.137\), \(0.933 \pm 0.052\), and \(0.870 \pm 0.095\). These results indicate that the pre-event \ac{EEG} window contains discriminative information across different traffic-event categories, and that \ac{3D-CRNN} can capture risk-related neural patterns more consistently than the compared baselines.

For \ac{DI}, \ac{3D-CRNN} with \ac{RSL} also achieves the best overall performance in all three event-wise evaluations. On \ac{AEB} scenarios, it obtains the highest F1-score of \(0.920 \pm 0.026\) and \ac{BA} of \(0.778 \pm 0.129\). On cut-in scenarios, it reaches the strongest performance, with an F1-score of \(0.892 \pm 0.014\) and \ac{BA} of \(0.880 \pm 0.026\). On pedestrian scenarios, it achieves an F1-score of \(0.924 \pm 0.018\) and \ac{BA} of \(0.799 \pm 0.074\). The relatively high F1-scores but lower \ac{BA} in \ac{AEB} and pedestrian scenarios suggest that these event-wise subsets may contain more asymmetric class-wise performance, making \ac{BA} a stricter and more informative metric.

The effect of \ac{RSL} is most consistent for the proposed \ac{3D-CRNN}. Compared with the corresponding setting without \ac{RSL}, \ac{3D-CRNN} with \ac{RSL} improves \ac{BA} from \(0.776\) to \(0.778\) on \ac{AEB}, from \(0.877\) to \(0.880\) on cut-in, and from \(0.760\) to \(0.799\) on pedestrian scenarios. The largest gain appears in pedestrian scenarios. One possible explanation is that pedestrians occupy a smaller visual area, making their hazardous behavior harder to observe and decode from the short post-event window alone. By introducing the earlier \ac{RP} signal, \ac{RSL} provides additional context about the passenger's pre-event vigilance toward the pedestrian, helping the model capture these subtle and less visually salient dangers. In contrast, several baseline models show mixed changes after \ac{RSL} in the event-wise setting, especially on \ac{AEB} and pedestrian scenarios. This indicates that simply combining \ac{RP} and \ac{DI} outputs is not always sufficient; the architecture must also effectively learn the temporal relationship between pre-event risk perception and near-onset danger identification.

\subsection{Cross-Session Generalizability in Danger Identification}
\label{res:cross-session}

Table~\ref{tab:cross_session_comparison} shows that all models experience lower performance in the \textit{cross-session} setting than in the within-session setting. Here, the held-out test session was strictly recorded on a different day, using the same protocol for every subject, making the evaluation more challenging due to intra-subject variability across sessions (e.g., mental state, recording conditions, and \ac{EEG} distribution shifts). Despite this degradation, the proposed \ac{3D-CRNN} still achieves the best overall performance, obtaining the highest Precision, Recall, F1-score, and \ac{BA}. In particular, it reaches a \ac{BA} of $0.770 \pm 0.053$, outperforming all compared baselines. These results indicate stronger robustness to session-related variability and better cross-session generalization for \ac{DI}.

\begin{table}[H]
\footnotesize
\centering
\setlength{\tabcolsep}{\dimexpr(\textwidth-13.3cm)/12\relax}
\begin{tabular}{
>{\centering\arraybackslash}p{2cm}
>{\centering\arraybackslash}p{2.5cm}
>{\centering\arraybackslash}p{2cm}
>{\centering\arraybackslash}p{2cm}
>{\centering\arraybackslash}p{2cm}
>{\centering\arraybackslash}p{2.8cm}
}
\toprule[1.5pt]
Task & Model & Precision $\uparrow$ & Recall $\uparrow$ & F1-score $\uparrow$ & Balanced Accuracy $\uparrow$ \\
\midrule
\multirow{7}{*}{Cross-session} & EEGNet & 0.717 $\pm$ 0.080 & 0.704 $\pm$ 0.042 & 0.690 $\pm$ 0.060 & 0.668 $\pm$ 0.097 \\
 & ShallowConvNet & \underline{0.763 $\pm$ 0.068} & \underline{0.732 $\pm$ 0.033} & \underline{0.733 $\pm$ 0.040} & \underline{0.731 $\pm$ 0.074} \\
 & DeepConvNet & 0.700 $\pm$ 0.085 & 0.664 $\pm$ 0.095 & 0.648 $\pm$ 0.101 & 0.643 $\pm$ 0.105 \\
 & XGB-DIM & 0.587 $\pm$ 0.021 & 0.555 $\pm$ 0.025 & 0.562 $\pm$ 0.023 & 0.541 $\pm$ 0.021 \\
 & CTNet & 0.687 $\pm$ 0.084 & 0.583 $\pm$ 0.094 & 0.576 $\pm$ 0.115 & 0.629 $\pm$ 0.082 \\
 & DSC-ConvLSTM & 0.635 $\pm$ 0.193 & 0.537 $\pm$ 0.112 & 0.510 $\pm$ 0.157 & 0.600 $\pm$ 0.102 \\
\rowcolor{AUINHighlight}
 & 3D-CRNN & \textbf{0.796 $\pm$ 0.056} & \textbf{0.747 $\pm$ 0.034} & \textbf{0.752 $\pm$ 0.034} & \textbf{0.770 $\pm$ 0.053} \\
\bottomrule
\end{tabular}

\vspace{2pt}
\caption{Cross-session performance comparison for Danger Identification.}
\label{tab:cross_session_comparison}
\end{table}

\subsection{Cross-Subject Generalizability in Danger Identification}
\label{res:cross-subject-di}
To evaluate the generalizability of our model, we conducted a cross-subject evaluation using the \ac{LOSO} cross-validation method. In each iteration, one subject was excluded from the seen group for testing, while the remaining seen-subject data were mixed and split into 50\% training, 20\% validation, and 30\% testing. The unseen-subject data were used exclusively for testing, and the final results were averaged across all iterations.

\begin{table}[H]
\footnotesize
\centering
\setlength{\tabcolsep}{\dimexpr(\textwidth-13.3cm)/12\relax}
\begin{tabular}{
>{\centering\arraybackslash}p{2cm}
>{\centering\arraybackslash}p{2.5cm}
>{\centering\arraybackslash}p{2cm}
>{\centering\arraybackslash}p{2cm}
>{\centering\arraybackslash}p{2cm}
>{\centering\arraybackslash}p{2.8cm}
}
\toprule[1.5pt]
Task & Model & Precision $\uparrow$ & Recall $\uparrow$ & F1-score $\uparrow$ & Balanced Accuracy $\uparrow$ \\
\midrule
\multirow{7}{*}{Seen-subject} & EEGNet & 0.798 $\pm$ 0.007 & \underline{0.800 $\pm$ 0.005} & \underline{0.798 $\pm$ 0.006} & 0.757 $\pm$ 0.012 \\
 & ShallowConvNet & 0.795 $\pm$ 0.008 & 0.798 $\pm$ 0.007 & 0.796 $\pm$ 0.008 & 0.751 $\pm$ 0.011 \\
 & DeepConvNet & 0.774 $\pm$ 0.016 & 0.780 $\pm$ 0.013 & 0.772 $\pm$ 0.023 & 0.713 $\pm$ 0.037 \\
 & XGB-DIM & 0.580 $\pm$ 0.007 & 0.503 $\pm$ 0.031 & 0.520 $\pm$ 0.031 & 0.501 $\pm$ 0.008 \\
 & CTNet & \underline{0.807 $\pm$ 0.047} & 0.736 $\pm$ 0.077 & 0.745 $\pm$ 0.078 & \underline{0.773 $\pm$ 0.064} \\
 & DSC-ConvLSTM & 0.764 $\pm$ 0.056 & 0.682 $\pm$ 0.124 & 0.682 $\pm$ 0.159 & 0.719 $\pm$ 0.083 \\
\rowcolor{AUINHighlight}
 & 3D-CRNN & \textbf{0.811 $\pm$ 0.008} & \textbf{0.812 $\pm$ 0.008} & \textbf{0.811 $\pm$ 0.008} & \textbf{0.774 $\pm$ 0.011} \\
\midrule
\multirow{7}{*}{Unseen-subject} & EEGNet & \underline{0.705 $\pm$ 0.061} & \underline{0.682 $\pm$ 0.055} & \underline{0.682 $\pm$ 0.052} & 0.644 $\pm$ 0.070 \\
 & ShallowConvNet & 0.673 $\pm$ 0.056 & 0.639 $\pm$ 0.108 & 0.631 $\pm$ 0.097 & 0.595 $\pm$ 0.071 \\
 & DeepConvNet & 0.660 $\pm$ 0.058 & 0.658 $\pm$ 0.087 & 0.650 $\pm$ 0.081 & 0.590 $\pm$ 0.070 \\
 & XGB-DIM & 0.588 $\pm$ 0.015 & 0.512 $\pm$ 0.030 & 0.531 $\pm$ 0.029 & 0.509 $\pm$ 0.016 \\
 & CTNet & 0.618 $\pm$ 0.109 & 0.451 $\pm$ 0.156 & 0.407 $\pm$ 0.209 & 0.558 $\pm$ 0.096 \\
 & DSC-ConvLSTM & 0.672 $\pm$ 0.207 & 0.602 $\pm$ 0.173 & 0.584 $\pm$ 0.228 & \textbf{0.661 $\pm$ 0.104} \\
\rowcolor{AUINHighlight}
 & 3D-CRNN & \textbf{0.713 $\pm$ 0.067} & \textbf{0.717 $\pm$ 0.064} & \textbf{0.709 $\pm$ 0.066} & \underline{0.649 $\pm$ 0.085} \\
\bottomrule
\end{tabular}

\vspace{2pt}
\caption{Cross-subject performance comparison on seen- and unseen-subject datasets for Danger Identification across held-out subjects.}
\label{tab:cross_subject_comparison}
\end{table}

Table~\ref{tab:cross_subject_comparison} presents the results for both seen-subject and unseen-subject evaluations. As expected, performance on unseen subjects is lower than on seen subjects, reflecting the challenge of inter-subject variability in \ac{EEG} decoding. In the seen-subject setting, \ac{3D-CRNN} achieves the best overall performance, with a Precision of \(0.811 \pm 0.008\), Recall of \(0.812 \pm 0.008\), F1-score of \(0.811 \pm 0.008\), and \ac{BA} of \(0.774 \pm 0.011\). CTNet obtains a very close second-best \ac{BA} of \(0.773 \pm 0.064\), but its Recall and F1-score are lower than those of \ac{3D-CRNN}.

In the unseen-subject setting, \ac{3D-CRNN} achieves the highest Precision, Recall, and F1-score, with values of \(0.713 \pm 0.067\), \(0.717 \pm 0.064\), and \(0.709 \pm 0.066\), respectively. Its \ac{BA} is \(0.649 \pm 0.085\), which is slightly lower than DSC-ConvLSTM's \(0.661 \pm 0.104\). Therefore, while \ac{3D-CRNN} provides the strongest overall classification consistency in terms of F1-score and class prediction quality, DSC-ConvLSTM shows a slight advantage in \ac{BA} for unseen subjects.

Overall, these results show that the proposed \ac{3D-CRNN} generalizes well compared with the tested baselines, especially in maintaining strong Precision, Recall, and F1-score under unseen-subject evaluation. Nevertheless, the performance gap between seen and unseen subjects remains clear, highlighting the difficulty of inter-subject variability in \ac{EEG}-based danger identification. Future work should further explore domain adaptation, subject calibration, or transfer learning strategies to improve robustness across different passengers.

\subsection{Model Complexity and Inference Latency}
\label{res:online}

To assess whether the prediction algorithm is suitable for real-time use, we benchmarked the proposed \ac{3D-CRNN} on Quadro RTX 6000 GPUs. The model contains 77,922 trainable parameters and requires 330.195\,M \acp{FLOP} per sample. Its average single-sample inference latency for the fused feature is 10.616\,ms, with 3.998\,ms and 3.945\,ms for \ac{RP} and \ac{DI} branches, respectively. These results suggest that the CRNN inference itself is compatible with real-time \ac{EEG} decoding, while the overall system latency will also depend on data acquisition and preprocessing.

\section{Discussion}
\label{sec:Discussion}

This study explored passenger \ac{EEG} as a risk-aware cognitive signal for highly automated vehicles. Unlike most existing \ac{EEG}-based driving studies that focus solely on immediate hazard detection, the proposed framework addresses both \ac{RP} and \ac{DI}, allowing potential risks to be captured before hazardous behavior fully unfolds.

A key contribution of this work is the \ac{PCM}, which links traffic conditions with passengers' neural responses and provides a neurocognitive basis for the decoding tasks. \ac{PCM} highlights two main transitions in passenger cognition: from \textit{Calm} to \textit{Nervous}, and from \textit{active observation} to \textit{decision-making}. Based on these transitions, we formulated \ac{RP} and \ac{DI}, and further introduced \ac{RSL} to reflect their sequential relationship, effectively integrating risk prediction signals to improve danger classification.

Building on \ac{PCM} and \ac{RSL}, the proposed \ac{PEDS} uses a \ac{3D-CRNN} to jointly model spatial topology and temporal dynamics in \ac{EEG}. The results show that passenger \ac{EEG} contains useful information for both anticipatory and event-related decoding. In particular, the proposed framework achieved a \ac{BA} of 0.953 for \ac{RP} and 0.850 for single-subject \ac{DI}. Event-wise analysis further showed that \ac{3D-CRNN} with \ac{RSL} achieved the best \ac{DI} performance across \ac{AEB}, cut-in, and pedestrian scenarios, with \ac{BA} values of 0.778, 0.880, and 0.799, respectively. In terms of generalizability, \ac{3D-CRNN} achieved the best performance in the cross-session setting, with a \ac{BA} of 0.770. In cross-subject evaluation, it achieved the highest Precision, Recall, and F1-score on both seen and unseen subjects, with the highest \ac{BA} of 0.774 on seen subjects and the second-highest \ac{BA} of 0.649 on unseen subjects. These findings support the feasibility of using passenger \ac{EEG} as an informative signal for \ac{AV} safety research and suggest that anticipatory neural activity can provide complementary information beyond immediate hazard responses.

\subsection{Limitations and Future Work}
\label{sec:limitations}

Several limitations and operational assumptions should be noted. First, the present framework was developed and evaluated under a controlled passenger-observation setting with a moderate number of valid participants. Participants were instructed to observe the simulated driving scene, and the auditory cue/button-press task was used to encourage task engagement. However, we do not assume that passengers remained fully attentive throughout the experiment. Since each participant completed multiple sessions and repeatedly observed similar traffic scenarios, fatigue, reduced vigilance, and habituation to repeated events may have weakened \ac{EEG} responses as the experiment progressed, contributing to intra-subject and cross-session variability. In addition, individual differences in age, sex, cognitive style, risk tolerance, prior driving experience, and neural response patterns may contribute to inter-subject variability and the performance drop on unseen subjects. Future work should include larger and more diverse cohorts and explicitly model attention levels, fatigue, scenario habituation, and subject-specific variability through domain adaptation, transfer learning, subject calibration, or multimodal indicators.

Second, the current \ac{RP}/\ac{DI} taxonomy focuses on observable ego--target traffic interactions, such as pedestrian crossing, vehicle cut-in, and front-vehicle emergency braking. Therefore, the proposed framework should be interpreted as an operational taxonomy for interaction-based hazards rather than a complete taxonomy of all AV hazards. Non-interactive hazards, such as occluded intersections, limited-visibility road segments, narrow passages, road-surface anomalies, or aggressive ego-vehicle maneuvers, may also elicit anticipatory \ac{EEG} responses through increased uncertainty, vigilance, or workload. Future work should extend the scenario set to these non-interactive hazards and examine whether their \ac{EEG} signatures follow the same risk-anticipation and danger-recognition pattern.

Third, the present study was conducted in a controlled simulator with a laboratory-grade wet \ac{EEG} system. While this setup provides relatively high signal quality and precise event-onset annotation, it does not fully reflect the signal conditions of real vehicles. In real driving environments, \ac{EEG} signals may be further contaminated by vehicle vibration, road-induced body movement, head motion, electrode displacement, cable motion, muscle activity, eye movement, and environmental electromagnetic noise. These factors may reduce the signal-to-noise ratio and weaken the stability of \ac{RP} and \ac{DI} decoding, especially for \ac{RP} because anticipatory \ac{EEG} responses can be more subtle than stimulus-locked post-onset responses. Future work should evaluate the framework in moving vehicles or high-fidelity motion simulators and develop artifact-robust preprocessing and modeling strategies, such as online artifact rejection, motion-aware filtering, simulator-to-real domain adaptation, and multimodal fusion with eye tracking, head motion, heart-rate variability, or vehicle dynamics.

Fourth, from an application perspective, \ac{RP} is particularly important because it provides an earlier opportunity for \ac{AV} systems to prepare or intervene before an immediate danger is instantiated. Although the proposed framework demonstrates promising \ac{RP} performance in the current interaction-based scenarios, future work should further improve the accuracy, robustness, and generalizability of \ac{RP} across broader driving conditions. In particular, future studies should investigate the trade-off between prediction lead time and decoding accuracy, since earlier \ac{RP} windows may provide more time for intervention but may also contain weaker or more ambiguous neural evidence.

Fifth, the inference latency of the proposed \ac{3D-CRNN} is 10.616\,ms, which is within a practical real-time range; however, the prediction algorithm is only one part of the full system pipeline. In real applications, end-to-end response time will also depend on \ac{EEG} acquisition, preprocessing, buffering, and system integration. Moreover, practical deployment would require more usable sensing hardware, such as portable or dry-electrode \ac{EEG} systems, whose robustness remains to be systematically validated. Future work should therefore evaluate the full online pipeline rather than the decoding model in isolation.

Finally, the use of neurophysiological data in vehicles raises important ethical and privacy considerations. Since \ac{EEG} may contain sensitive information about cognitive or affective states, any future deployment would require clear safeguards for informed consent, secure data handling, controlled access, and transparent data governance. While these issues are beyond the scope of the present technical study, they are important for responsible translation.

Overall, the present work should be viewed as a controlled proof-of-concept study. The results demonstrate the promise of passenger \ac{EEG} for \ac{AV} safety research, but they do not yet establish deployment-ready performance across users and real-world conditions. Future work will focus on improving \ac{RP} robustness, expanding the dataset, validating the framework in more realistic driving environments, and integrating \ac{EEG} with other physiological or behavioral signals.

\section{Conclusion}
\label{sec:Conclusion}

This paper presented a passenger-centered \ac{EEG} decoding framework for highly automated vehicles, targeting both \acf{RP} and \acf{DI}. To this end, we introduced the \ac{PCM}, \ac{RSL}, and the \ac{PEDS} based on a \ac{3D-CRNN} architecture. Together, these components provide a unified framework that links passenger neural mechanisms, task formulation, and \ac{EEG} decoding.

Experimental results showed that the proposed framework can effectively decode passenger cognitive responses to traffic risk, achieving a \ac{BA} of 0.953 for \ac{RP} and a \ac{BA} of 0.850 for single-subject \ac{DI}, while also achieving the highest cross-subject Precision, Recall, and F1-score among the evaluated baselines. These results indicate that passenger \ac{EEG} contains both anticipatory and hazard-related information that can support \ac{AV} safety research beyond immediate hazard recognition alone.

At the same time, the present work remains a controlled proof-of-concept study, with limitations including the moderate sample size, the simulator-based setting, and the remaining challenge of inter-subject variability. Future work should therefore focus on larger and more diverse cohorts, stronger subject-independent modeling, more realistic online validation, and multimodal approaches that combine \ac{EEG} with other physiological or behavioral signals.

Overall, this study shows that passenger \ac{EEG} is a meaningful source of cognitive information that can inform the development, training, and evaluation of safer and more adaptive autonomous vehicles.

\section*{Acknowledgements}
We would like to express our sincere gratitude to all individuals and institutions that supported this work. The authors affiliated with the School of Vehicle and Mobility, Tsinghua University, and Yingkai Yang acknowledge support for this research from the National Science Foundation of China Project 52072215, National Natural Science Foundation of China grant 52221005, National Natural Science Foundation of China grant 52075213, Beijing Natural Science Foundation grant L243025, National Key R\&D Program of China grant 2022YFB2503003, and the State Key Laboratory of Intelligent Green Vehicle and Mobility. We extend our thanks to all the participants who contributed to the data collection process. We would also like to acknowledge our colleagues and collaborators for their valuable discussions and technical assistance throughout the research.

\section*{Ethics Statement}
This study complied with the Declaration of Helsinki and was approved by the Institutional Review Board of Tsinghua University, China. Participants were informed of their right to ask questions about the study and were assured that their personal and identifiable data would remain confidential. Additionally, they were encouraged to take breaks whenever necessary and were free to end the session at any time for any reason. All individuals provided written informed consent after receiving a full explanation of the study procedures. Participants were also compensated for their participation.

\section*{Author Contributions}
Yingkai Yang and Ashton Yu Xuan Tan prepared the manuscript; Yingkai Yang, Ashton Yu Xuan Tan, Xiaorong Gao, Hong Wang, and Bowen Li handled the data and completed the data analysis; Jun Li, Hong Wang, Jianqiang Wang, and Sifa Zheng provided financial support; Yingkai Yang and Ashton Yu Xuan Tan designed the model and algorithms and conducted the experiments. Xinyu Gu, Yang Zhao, Yuxin Zhang, Sharon X. Huang, and Tania Stathaki revised the manuscript.

\section*{Conflict of Interest}
On behalf of all authors, the corresponding author states that there are no conflicts of interest.

\bibliographystyle{unsrtnat}
\bibliography{sn-bibliography}

\clearpage
\appendix
\input{supplementary}

\end{document}

%% file: supplementary.tex
\section{Scenario Specifications}
\label{app:scenario-specifications}

\begin{figure}[H]
     \centering
         \includegraphics[width=\textwidth]{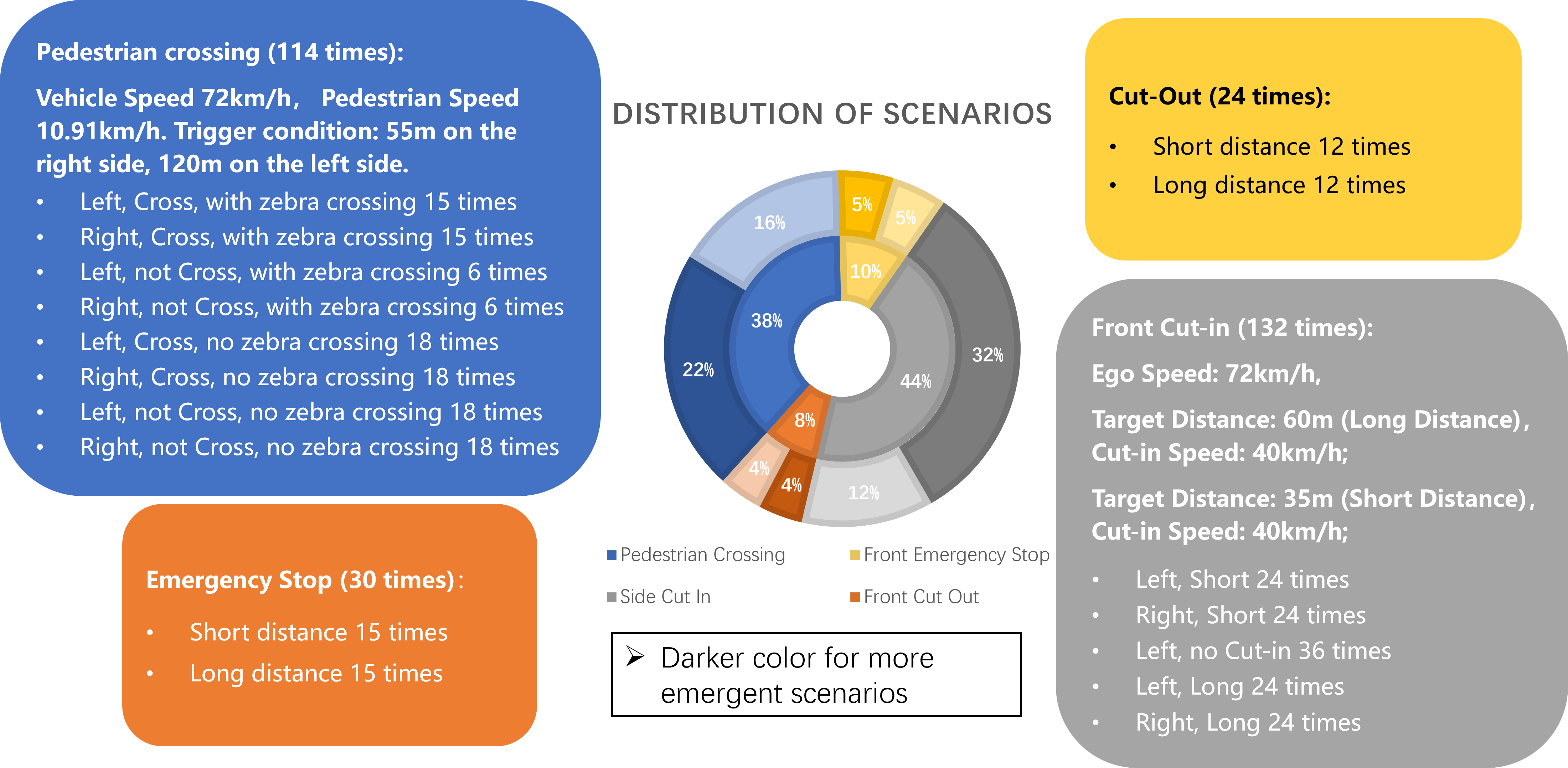}
        \caption{Scenario Specifications and Distribution.}
        \label{fig:event-distribution}
\end{figure}

The 14 traffic scenarios cover pedestrian encounters, vehicle cut-ins, and lead-vehicle cut-outs with or without emergency braking. Table~\ref{tab:scenario-examples} summarizes their clip counts, risk levels, and example views, while Fig.~\ref{fig:event-distribution} shows the scenario settings and event distribution.

\begin{table}[H]
\centering
\small
\setlength{\tabcolsep}{4pt}
\renewcommand{\tabularxcolumn}[1]{m{#1}}
\begin{tabularx}{\textwidth}{@{}
    >{\centering\arraybackslash}m{0.05\textwidth}
    >{\raggedright\arraybackslash}X
    >{\centering\arraybackslash}m{0.14\textwidth}
    >{\centering\arraybackslash}m{0.10\textwidth}
    >{\centering\arraybackslash}m{0.33\textwidth}@{}}
\toprule
    \textbf{ID} & \textbf{Description} & \shortstack{\textbf{Number}\\\textbf{of Clips}} & \shortstack{\textbf{Risk}\\\textbf{Level}} & \textbf{Example} \\
\midrule
\shortstack{\strut 1\\\strut 3} &
\shortstack[l]{\strut Pedestrian Cross Left\\\strut Pedestrian Stand Left} &
\shortstack{\strut 33\\\strut 24} &
\shortstack{\strut High\\\strut Low} &
\includegraphics[width=\linewidth,height=0.14\textheight,keepaspectratio]{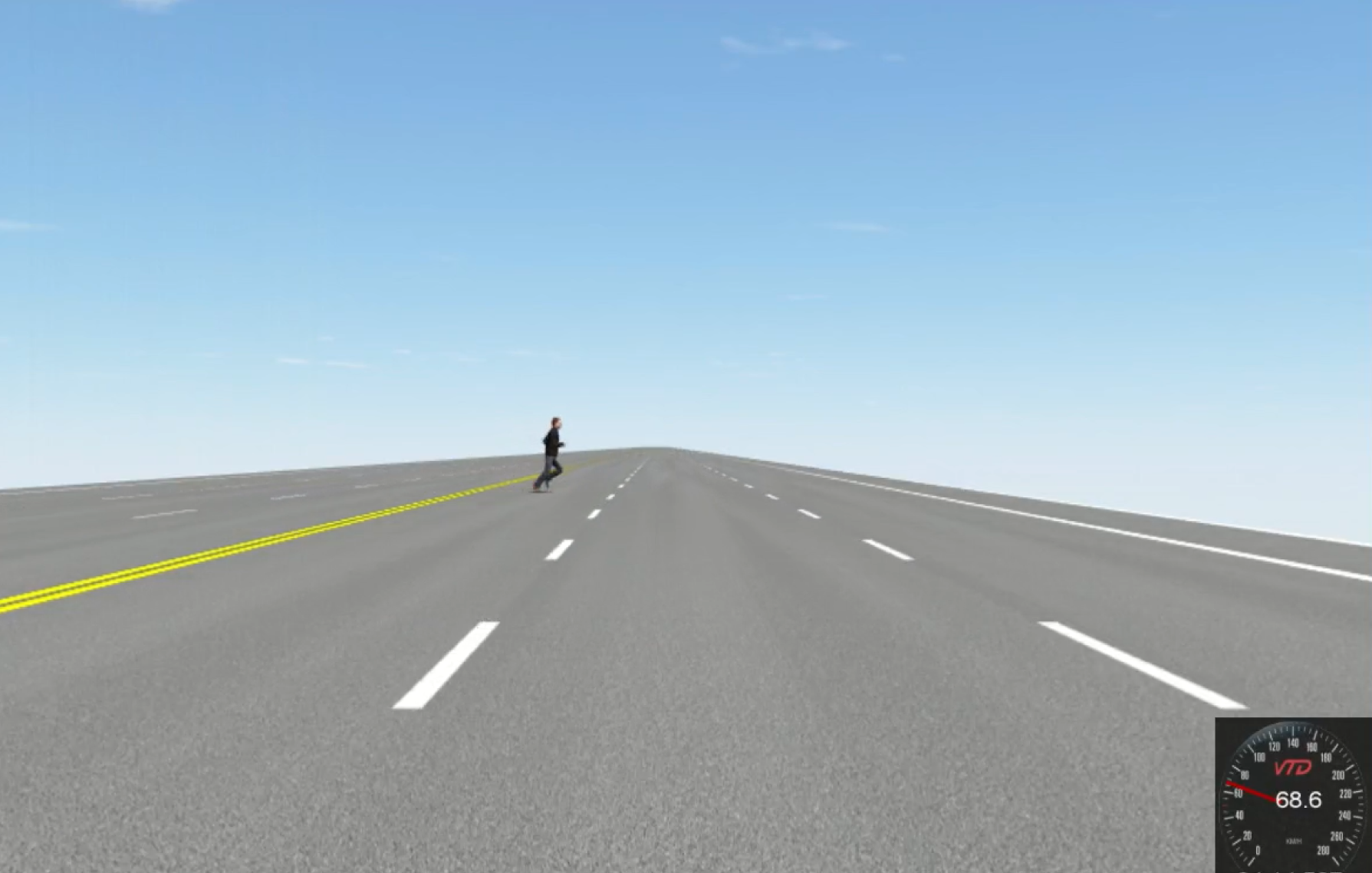} \\
\midrule
\shortstack{\strut 2\\\strut 4} &
\shortstack[l]{\strut Pedestrian Cross Right\\\strut Pedestrian Stand Right} &
\shortstack{\strut 33\\\strut 24} &
\shortstack{\strut High\\\strut Low} &
\includegraphics[width=\linewidth,height=0.14\textheight,keepaspectratio]{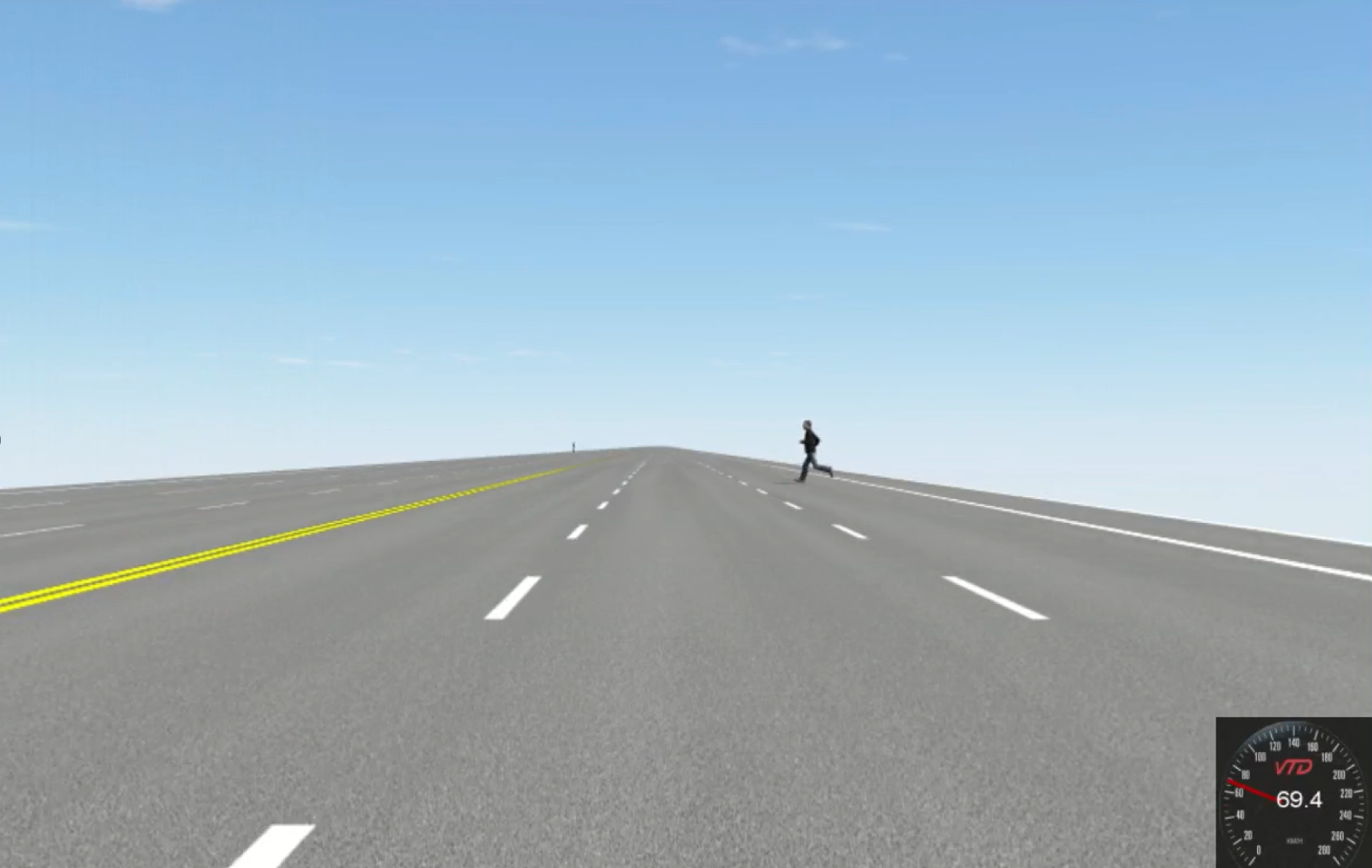} \\
\midrule
\shortstack{\strut 5\\\strut 7\\\strut 9} &
\shortstack[l]{\strut Vehicle Non-Cut-in Left\\\strut Vehicle Cut-in Left Close\\\strut Vehicle Cut-in Left Far} &
\shortstack{\strut 18\\\strut 24\\\strut 24} &
\shortstack{\strut Low\\\strut High\\\strut High} &
\includegraphics[width=\linewidth,height=0.14\textheight,keepaspectratio]{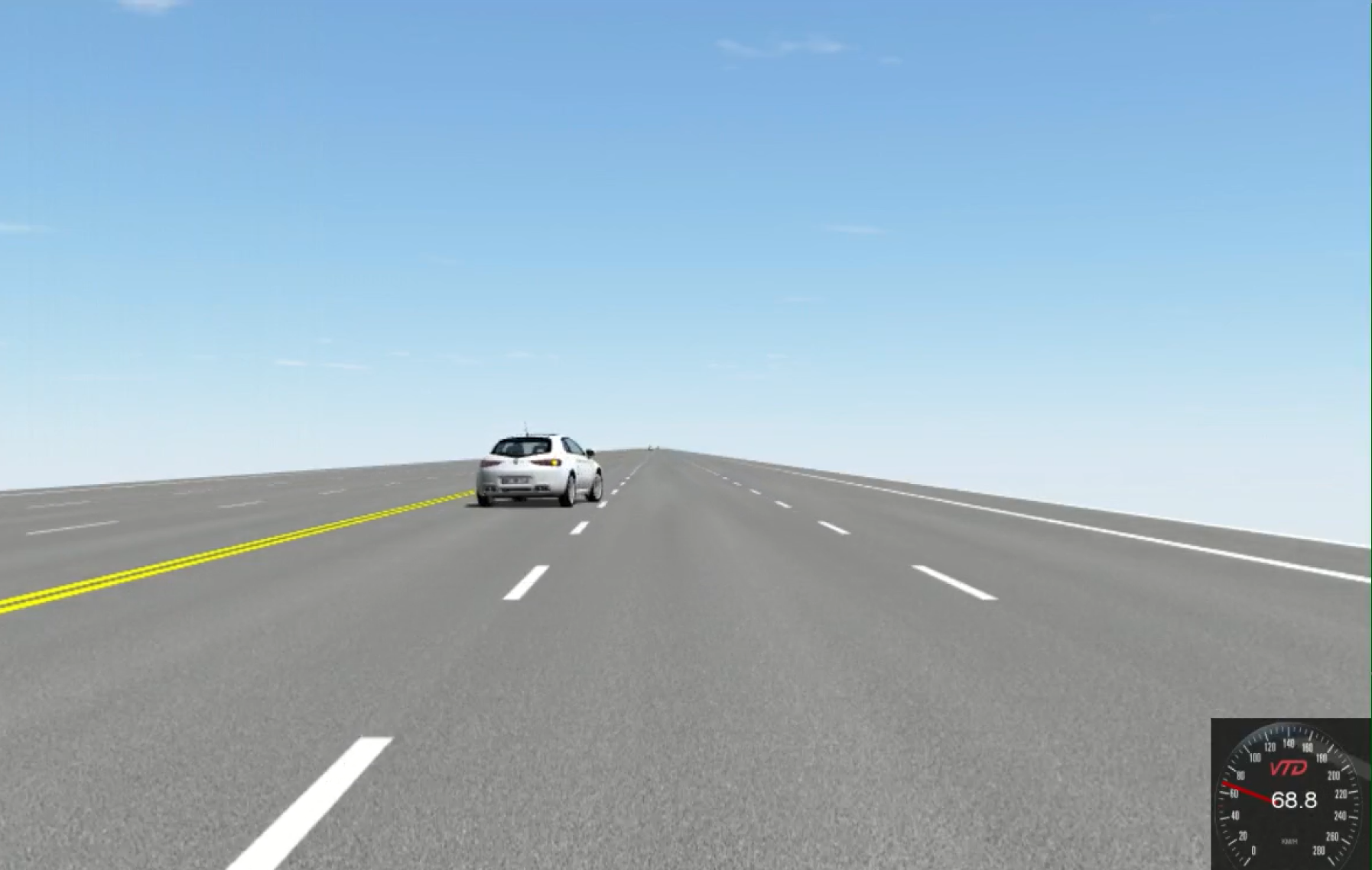} \\
\midrule
\shortstack{\strut 6\\\strut 8\\\strut 10} &
\shortstack[l]{\strut Vehicle Non-Cut-in Right\\\strut Vehicle Cut-in Right Close\\\strut Vehicle Cut-in Right Far} &
\shortstack{\strut 18\\\strut 24\\\strut 24} &
\shortstack{\strut Low\\\strut High\\\strut High} &
\includegraphics[width=\linewidth,height=0.14\textheight,keepaspectratio]{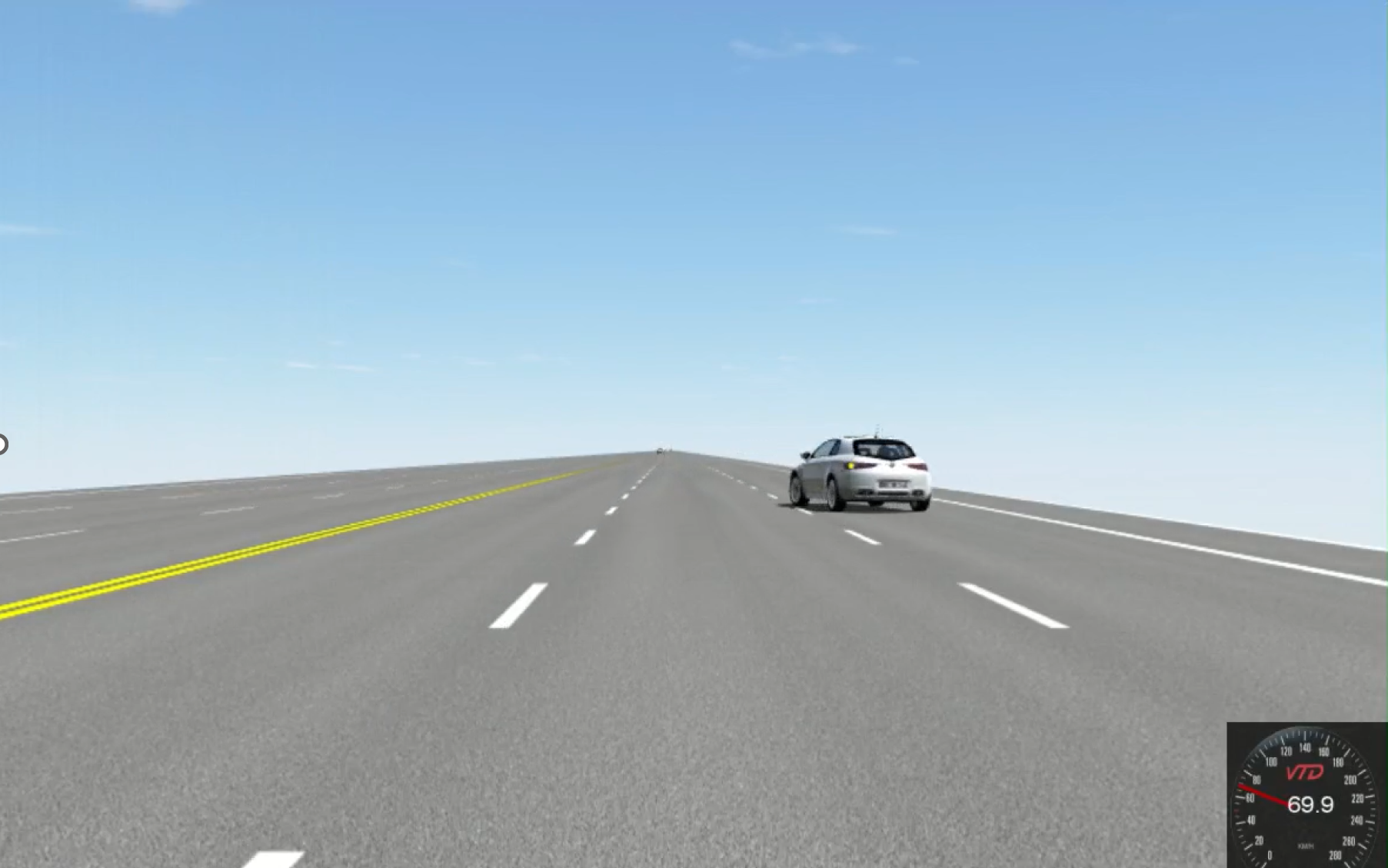} \\
\midrule
\shortstack{\strut 11\\\strut 12\\\strut 13\\\strut 14} &
\shortstack[l]{\strut Vehicle Cut-out Close\\\strut Vehicle Cut-out Far\\\strut Vehicle AEB Close\\\strut Vehicle AEB Far} &
\shortstack{\strut 12\\\strut 12\\\strut 15\\\strut 15} &
\shortstack{\strut Low\\\strut Low\\\strut High\\\strut High} &
\includegraphics[width=\linewidth,height=0.14\textheight,keepaspectratio]{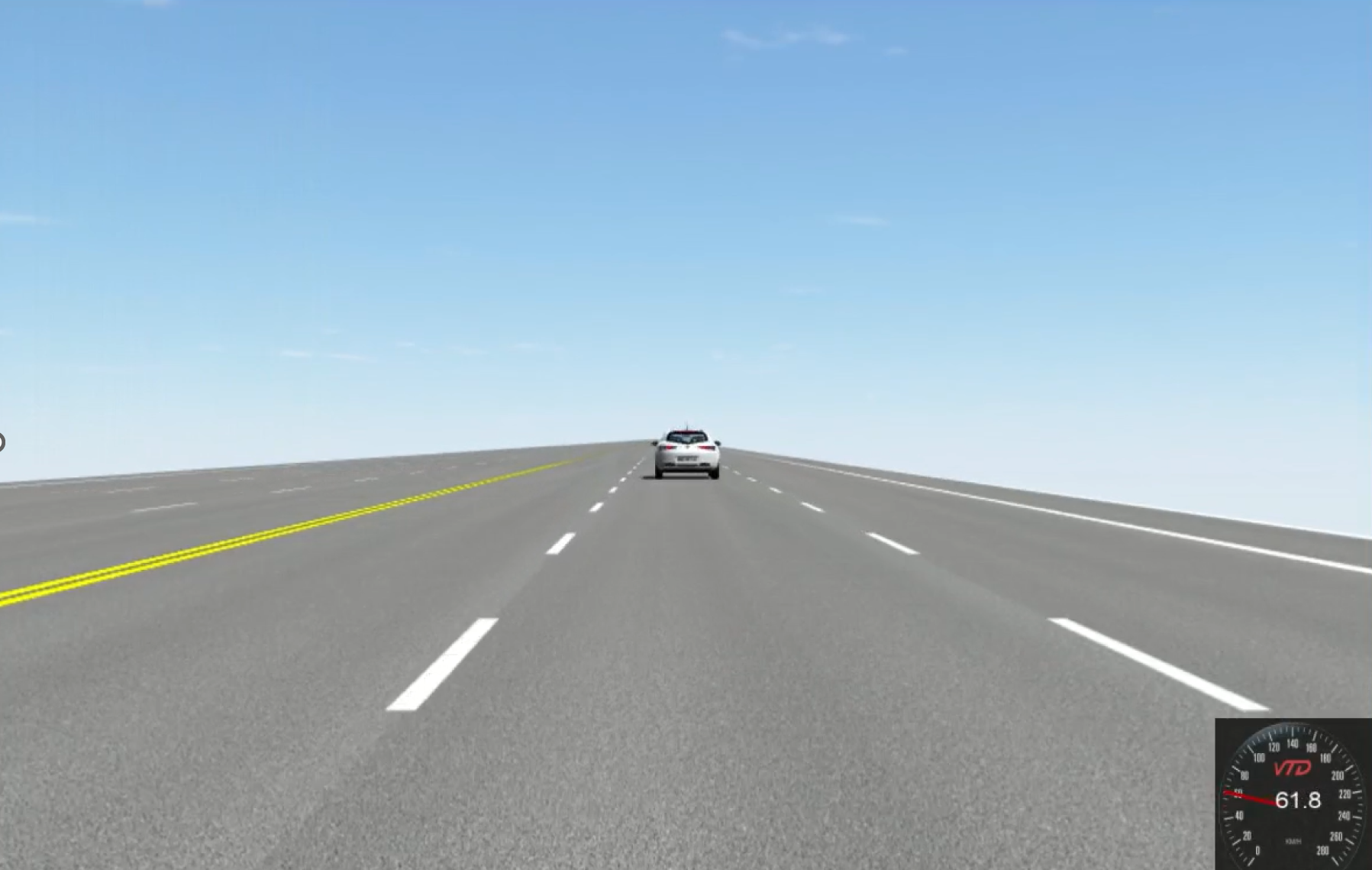} \\
\bottomrule
\end{tabularx}
\caption{Scenario Settings and Examples: Each row lists the scenario ID, a short description, the number of clips per scenario, the risk level, and an example visualization of the traffic scenario in VTD.}
\label{tab:scenario-examples}
\end{table}

\section{Data Split and Composition}
\label{app:data-split}

To make the evaluation protocol explicit, we report the composition of all training, validation, and test partitions used in this study. The 11 valid subjects had subject IDs 1--8 and 10--12. For the single-subject and event-wise protocols, data from sessions 1--4 were first split per class into training and testing sets with a 70:30 ratio. The training partition was then shuffled with \texttt{random\_state=100} and further divided into training and validation sets using \texttt{validation\_split=0.3}. For the cross-session protocol, sessions 1--3 were used for training/validation and session 4 was reserved for testing. For the cross-subject protocol, we used leave-one-subject-out (LOSO) validation: in each fold, one subject was held out as the unseen test subject, while training, validation, and seen testing samples were drawn only from the remaining ten seen subjects. For Risk Prediction (RP), ``Risk'' includes both low-risk and high-risk samples, while ``Safe'' corresponds to non-risk samples. For Danger Identification (DI), ``Danger'' corresponds to high-risk samples, while ``Non-danger'' includes safe and low-risk samples.

\begin{table}[H]
\centering
\small
\setlength{\tabcolsep}{6pt}
\renewcommand{\arraystretch}{1.15}
\begin{tabular*}{\textwidth}{@{\extracolsep{\fill}}ccccccc@{}}
\toprule
\textbf{Split} & \textbf{Samples} & \shortstack{\textbf{Samples/}\\\textbf{Subj.}} & \textbf{Safe} & \textbf{Risk} & \textbf{Non-danger} & \textbf{Danger} \\
\midrule
Train      & 3964 & 360 & 2056 & 1908 & 2765 & 1199 \\
Validation & 1707 & 155 & 1024 & 683  & 1202 & 505  \\
Test       & 2444 & 222 & 1320 & 1124 & 1708 & 736  \\
\bottomrule
\end{tabular*}
\vspace{1mm}

{\small \emph{Note:} This split uses 11 subjects, sessions 1--4, and all event scenarios, including AEB, cut-in, and left- and right-side pedestrian events. RP labels are Safe/Risk; DI labels are Non-danger/Danger.}
\caption{Composition of the all-event single-subject data split.}
\label{tab:data_split_all_event}
\end{table}

\begin{table}[H]
\centering
\small
\setlength{\tabcolsep}{5pt}
\renewcommand{\arraystretch}{1.15}
\begin{tabular*}{\textwidth}{@{\extracolsep{\fill}}cccccccc@{}}
\toprule
\textbf{Scenario} & \textbf{Split} & \textbf{Samples} & \shortstack{\textbf{Samples/}\\\textbf{Subj.}} & \textbf{Safe} & \textbf{Risk} & \textbf{Non-danger} & \textbf{Danger} \\
\midrule
AEB & Train      & 2548 & 232 & 2156 & 392 & 2351 & 197 \\
AEB & Validation & 1101 & 100 & 924  & 177 & 991  & 110 \\
AEB & Test       & 1579 & 144 & 1320 & 259 & 1441 & 138 \\
\midrule
Cut-in & Train      & 3119 & 284 & 2125 & 994 & 2407 & 712 \\
Cut-in & Validation & 1340 & 122 & 955  & 385 & 1039 & 301 \\
Cut-in & Test       & 1926 & 175 & 1320 & 606 & 1484 & 442 \\
\midrule
Pedestrian & Train      & 5132 & 467 & 4285 & 847 & 4681 & 451 \\
Pedestrian & Validation & 2220 & 202 & 1875 & 345 & 1993 & 227 \\
Pedestrian & Test       & 3178 & 289 & 2640 & 538 & 2871 & 307 \\
\bottomrule
\end{tabular*}
\vspace{1mm}

{\small \emph{Note:} All rows use 11 subjects and sessions 1--4. The pedestrian scenario combines left- and right-side pedestrian events.}
\caption{Scenario-wise composition of the event-wise data splits.}
\label{tab:data_split_event_wise}
\end{table}

\begin{table}[H]
\centering
\small
\setlength{\tabcolsep}{2pt}
\renewcommand{\arraystretch}{1.15}
\begin{tabular*}{\textwidth}{@{\extracolsep{\fill}}ccccccccc@{}}
\toprule
\textbf{Split} & \textbf{Subjects} & \textbf{Sessions} & \textbf{Samples} & \shortstack{\textbf{Samples/}\\\textbf{Subj.}} & \textbf{Safe} & \textbf{Risk} & \textbf{Non-danger} & \textbf{Danger} \\
\midrule
Train      & 11 & 1--3 & 4421 & 402 & 2275 & 2146 & 3060 & 1361 \\
Validation & 11 & 1--3 & 1902 & 173 & 1025 & 877  & 1272 & 630  \\
Test       & 11 & 4    & 2620 & 238 & 1100 & 1520 & 1726 & 894  \\
\bottomrule
\end{tabular*}
\vspace{1mm}

{\small \emph{Note:} This split uses all event scenarios, including AEB, cut-in, and left- and right-side pedestrian events.}
\caption{Composition of the cross-session data split.}
\label{tab:data_split_cross_session}
\end{table}

\begin{table}[H]
\centering
\small
\setlength{\tabcolsep}{3pt}
\renewcommand{\arraystretch}{1.15}
\begin{tabular*}{\textwidth}{@{\extracolsep{\fill}}ccccccccc@{}}
\toprule
\textbf{Role} & \textbf{Folds} & \shortstack{\textbf{Subj./}\\\textbf{Fold}} & \shortstack{\textbf{Total}\\\textbf{Samples}} & \shortstack{\textbf{Samples/}\\\textbf{Fold}} & \textbf{Safe} & \textbf{Risk} & \textbf{Non-danger} & \textbf{Danger} \\
\midrule
Train       & 11 & 10 & 39691 & 3608 & 1949 & 1659 & 2507 & 1101 \\
Validation  & 11 & 10 & 17019 & 1547 & 851  & 696  & 1100 & 448  \\
Seen test   & 11 & 10 & 24440 & 2222 & 1200 & 1022 & 1553 & 669  \\
Unseen test & 11 & 1  & 8115  & 738  & 400  & 338  & 516  & 222  \\
\bottomrule
\end{tabular*}
\vspace{1mm}

{\small \emph{Note:} This split uses all event scenarios. In LOSO evaluation, ``Seen test'' contains samples from the ten subjects included in the training/validation pool, whereas ``Unseen test'' contains all samples from the held-out subject. Fold-level averages are rounded to the nearest integer.}
\caption{Composition of the cross-subject LOSO data split.}
\label{tab:data_split_cross_subject}
\end{table}